\documentclass{article}

\usepackage{microtype}
\usepackage{graphicx}
\usepackage{caption, subcaption}  
\usepackage{booktabs} 
\usepackage{multirow}
\usepackage{array} 
\usepackage[most]{tcolorbox}
\usepackage{pifont, arydshln}
\usepackage{hyperref}
 \usepackage{numprint}
\newcommand{\graybox}[1]{\textcolor{magenta}{#1}}

\newcommand{\cmark}{\ding{51}}%
\newcommand{\xmark}{\ding{55}}%
\usepackage[accepted]{icml2024_1}

\usepackage{amsmath}
\usepackage{amssymb}
\usepackage{mathtools}
\usepackage{amsthm}
\usepackage{enumitem}
\usepackage[capitalize,noabbrev]{cleveref}

\usepackage{xcolor}
\usepackage[normalem]{ulem} 
\usepackage{blindtext}%
\usepackage[textsize=tiny]{todonotes}
\presetkeys{todonotes}{fancyline}{}

\theoremstyle{plain}

\theoremstyle{definition}

\theoremstyle{remark}

\newcommand{\task}{VISLA }
\newcommand{\tasks}{VISLA}

\icmltitlerunning{VISLA Benchmark}

\begin{document}

\twocolumn[
\icmltitle{VISLA Benchmark: Evaluating Embedding Sensitivity to \\Semantic and Lexical Alterations}
\icmlsetsymbol{equal}{*}

\begin{icmlauthorlist}
\icmlauthor{Sri Harsha Dumpala}{equal,dal,vec}
\icmlauthor{Aman Jaiswal}{equal,dal}
\icmlauthor{Chandramouli Sastry}{dal,vec}
\icmlauthor{Evangelos Milios}{dal}
\icmlauthor{Sageev Oore}{dal,vec}
\icmlauthor{Hassan Sajjad}{dal}
\end{icmlauthorlist}

\icmlaffiliation{dal}{Dalhousie University, Canada}
\icmlaffiliation{vec}{Vector Institute, Canada}
\icmlcorrespondingauthor{Sri Harsha Dumpala}{sriharsha.d@dal.ca}
\icmlkeywords{Machine Learning, ICML}

\vskip 0.3in
]



\printAffiliationsAndNotice{\icmlEqualContribution} 

\begin{abstract}

Despite their remarkable successes, state-of-the-art language models face challenges in grasping certain important semantic details. This paper introduces the VISLA (Variance and Invariance to Semantic and Lexical Alterations) benchmark, designed to evaluate the semantic and lexical understanding of language models. VISLA presents a 3-way semantic (in)equivalence task with a triplet of sentences associated with an image, to evaluate both vision-language models (VLMs) and unimodal language models (ULMs). An evaluation involving 34 VLMs and 20 ULMs reveals surprising difficulties in distinguishing between lexical and semantic variations. Spatial semantics encoded by language models also appear to be highly sensitive to lexical information. Notably, text encoders of VLMs demonstrate greater sensitivity to semantic and lexical variations than unimodal text encoders.
Our contributions include the unification of image-to-text and text-to-text retrieval tasks, an off-the-shelf evaluation without fine-tuning, and assessing LMs' semantic (in)variance in the presence of lexical alterations. The results highlight strengths and weaknesses across diverse vision and unimodal language models, contributing to a deeper understanding of their capabilities. 
Data and code will be made available at \url{https://github.com/Sri-Harsha/visla_benchmark/}.
\end{abstract}

\section{Introduction}

Embeddings derived from state-of-the-art large language models (LLMs) form the foundation of several downstream applications, and even achieve human-level performance for some tasks~\cite{zhou2023large}. Despite such success, LLMs are limited in their precise understanding of the semantics of the language. For instance, they exhibit different behaviors for semantically equivalent sentences composed with different syntactic/lexical structures  \citep{krishna2023paraphrasing, p0, p1, p2}.  These challenges persist in vision-language models (VLMs) such as CLIP \cite{radford2021learning}, particularly in the form of visio-linguistic compositionality -- the difficulty in matching images to text describing their composition  \cite{thrush2022winoground, yuksekgonul2023and}. Specifically, the image representation from VLMs is found to be more biased towards matching the word(s) from the text rather than the semantics derived from their composition.  

\begin{figure}[ht!]
\hspace{-0.2cm}
    \centering
    \includegraphics[width=0.75\linewidth]{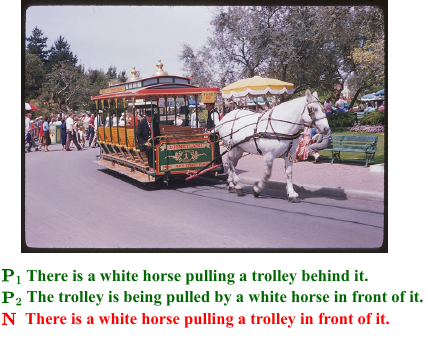}
    \vspace{-8pt}
    \caption{\small Figure shows an example from our VISLA benchmark. $P_1$ and $P_2$ are semantically equivalent but lexically different while $N$ is semantically different than both $P_1$ and $P_2$ despite its lexical similarity with $P_1$. In our evaluations of state-of-the-art language models (consisting of 34 VLMs and 20 ULMs) on this example, we (surprisingly) find that none of them are able to successfully identify the semantically equivalent pair ($P_1$, $P_2$) from the semantically different pairs (($P_1$, $N$), ($P_2$, $N$)).} 
    \label{fig:example}
    \vspace{-8pt}
\end{figure}

Despite prior attempts to assess VLMs in visio-linguistic compositional reasoning, it remains unclear if the information gap lies in the image or text encoder. Further, the ambiguity persists regarding whether the text representation is enriched with compositional semantic information and how invariant this representation might be with respect to the lexical choices used to convey the semantics. For example, 
Figure~\ref{fig:example} presents an example image with three captions (${\bf P}_1, {\bf P}_2, {\bf N}$). Let's consider the first caption (${\bf P}_1$) as our point of comparison. Semantically, ${\bf P}_1$ and ${\bf P}_2$ are equivalent, and, while ${\bf P}_1$ and the third caption (${\bf N}$) are semantically opposite, they are syntactically and lexically similar. \textit{Can LLMs resolve these peculiar differences between lexical similarity and semantic similarity?} In other words, do they understand the semantic relationships between the three sentences beyond the syntactic form?

In this work, we 
systematically evaluate language models' understanding of semantic and lexical differences between input text. We develop a benchmark dataset \emph{VISLA}, \textbf{V}ariance and \textbf{I}nvariance to \textbf{S}emantic and \textbf{L}exical \textbf{A}lterations, to achieve this\footnote{We will make this dataset publicly available upon acceptance of this work.}. 
The intuition of VISLA is to disentangle the semantic and lexical similarities when interpreting the representational capabilities of a language model. VISLA achieves this by defining a set of three related captions for an image:
${\bf P}_1$, a caption of the image;
${\bf P}_2$, another caption, which is semantically equivalent to ${\bf P}_1$ but lexically different; and
${\bf N}$, an incorrect caption of the image which is lexically close to ${\bf P}_1$ but semantically opposite to both ${\bf P}_1$ and ${\bf P}_2$. This is also referred to as a \textit{hard} negative caption \citep{hsieh2023sugarcrepe}.
This triplet allows us to evaluate the compositional capability of VLMs and LLMs, while disentangling semantic and lexical similarities. Unlike the STS benchmark(s)~\cite{sts17t1} that evaluates the degree of semantic similarity between pairs of text snippets, \task is a 3-way semantic (in)equivalence task across varying levels of lexical shifts. Specifically, the two positives and a hard negative eliminate the trivial case of selecting between a positive and a negative as is usually done in previous benchmarks(e.g., ~\cite{hsieh2023sugarcrepe}).

VISLA offers two benchmarking datasets: (a) generic, and (b) spatial. The generic dataset evaluates a model's ability to understand equivalent semantics with lexical variations, while the spatial dataset examines the ability of language models to correctly identify sentences describing similar spatial arrangements. In the \task triplets, semantically equivalent text pairs are visually represented with an image (refer to Figure \ref{fig:example}). This design allows us to evaluate VLMs in both multimodal (vision-language) and unimodal (text-only) settings and compare their performance with LLMs, which we refer to as Unimodal Language Models (ULMs).

We consider an embedding-based methodology to evaluate language models using VISLA. The embeddings of a language model represent how it encodes semantic and lexical knowledge. We hypothesize that for an LM to correctly understand the relation between positive and negative sentences, it must encode the positive sentences closer to each other than either one to the hard negative sentence. Based on this, we perform an extensive evaluation of 34 VLMs and 20 ULMs in their ability to understand variance in semantic and lexical alteration. A few of the notable findings are summarized below:
\begin{itemize}
    \item All text encoders---irrespective of their architecture, model size, training data size and optimization objective---struggle to separate out lexical and semantic variations.
    \item Spatial understanding in language models is highly sensitive to lexical information, and lexical overlap can divert models from capturing spatial semantics.
    \item Text encoders of vision LMs are more sensitive to semantic and lexical variations than unimodal text encoders.
    \item The pretraining of vision LMs that aligns text with semantic concepts of images showed better semantic understanding compared to CLIP.
\end{itemize}

Our contributions are as follows:
\begin{enumerate}
    \item The \task task unifies image-to-text retrieval (VLMs) and text-to-text retrieval (VLMs, ULMs), enabling the evaluation and comparison of both VLMs and ULMs within a single benchmark.
    
    

    \item The \task benchmark entails a 3-way semantic (in)equivalence task with varying lexical shifts, offering a more rigorous evaluation compared to benchmarks featuring only semantically distant text pairs. The hard negative caption ($\textbf{N}$) enables the assessment of semantic (in)variance of representation in the presence of lexical variance.
    
    \item We perform a thorough evaluation of a large set of vision and unimodal language models, 
    highlighting their strengths and weaknesses.
\end{enumerate}

\section{Related Work}

VLMs and ULMs have achieved impressive results on a range of vision and language downstream tasks. These state-of-the-art VLMs and ULMs serve as foundation models for both multimodal applications, like image captioning \cite{li2023blip2}, semantic segmentation of images \cite{Ding0XD22, LiangWDLZ0ZVM23}, text-to-image generation \cite{dalle, dalle-2, imagen}, and unimodal applications, like clustering \cite{wang2023improving, wang2022text}, reranking \cite{bge_embedding}, and retrieval \cite{li2023angle}. 
Their emergence as foundation models has motivated recent research to evaluate the strengths and weaknesses of these models. We summarize the findings from common benchmarks of VLMs and ULMs below.

\vspace{-5pt}
\paragraph{Findings from the Existing Benchmark for VLMs:}
 \citet{thrush2022winoground} evaluate VLMs through an image-text retrieval task and find that SOTA VLMs struggle to distinguish between texts containing the same words but ordered differently. Similarly, \citet{yuksekgonul2023and} evaluate VLMs in terms of their abilities to form object-attribute associations and highlight shortcomings of VLMs. Other studies with similar conclusions include \cite{zhao2022vl}, \cite{ray2023cola} and \cite{wang2023can}.
Recent works have introduced benchmarks to evaluate different abilities of VLMs such as counter-intuitive reasoning \cite{rome_dataset}, visual question answering \cite{xu2023lvlm}, conceptual understanding \cite{schiappa2023probing}, visio-linguistic reasoning \cite{chow2023travlr}, visual-spatial reasoning \cite{liu2023visual} and compositionality \cite{thrush2022winoground}. \citet{kamath2023text} demonstrate challenges in decoding salient aspects of input text encoded with CLIP and draw connections to the lack of compositionality in CLIP text embeddings. The task of evaluating compositionality in VLMs is the nearest neighbor to our work. Several datasets have been introduced to evaluate the compositionality of VLMs \cite{liu2023visual, hsieh2023sugarcrepe, thrush2022winoground, zhao2022vl, yuksekgonul2023and, ma2023crepe, ray2023cola, wang2023can, sahin2024enhancing}. Most of the existing compositionality benchmarks formulate the evaluation task as image-text retrieval. Winoground \cite{thrush2022winoground} is one of the earliest benchmarks to report the lack of compositional understanding in VLMs. Latest benchmarks encompassing different aspects of compositionality include VL-CheckList~\cite{zhao2022vl}, CREPE~\cite{ma2023crepe}, Cola~\cite{ray2023cola}, and ARO~\cite{yuksekgonul2023and}. Some benchmarks like Winoground have challenges beyond compositionality that include additional visual and textual reasoning \cite{DiwanBCHM22}. 




\vspace{-5pt}
\paragraph{Findings from the Existing Benchmarks for ULMs:}
In the context of ULM text encoders, paraphrasing is the closest to our \task task. Paraphrasing is a well-studied problem in NLP. Several previous studies analyzed the ability of the language models to recognize paraphrasing in text. The Microsoft Research Paraphrase Corpus (MRPC) \cite{dolan2005automatically} and Quora Question Pairs (QQP) \cite{qqp_link} are popular paraphrasing datasets (text-only without images) that are part of the GLUE (General Language Understanding Evaluation) \cite{WangSMHLB19} benchmark. The Semantic Textual Similarity (STS) benchmark \cite{sts17t1} build from the STS shared tasks \cite{sts12t6,sts13,sts14t10,sts15t2,sts16t1} have pairs of text snippets with scores indicating the degree of semantic equivalence between them. 

`~

\vspace{-5pt}
\paragraph{Shortcoming of existing Benchmarks:}
\citet{alper2023bert} find that the CLIP text encoder outperforms the ULMs in tasks that require implicit visual reasoning, while \citet{ChenCDWW23} find that ULMs perform better in terms of general language understanding. The ambiguities in the findings enforce the requirement of a more stringent benchmark with a precise diagnostic ability to discern the weaker encoder among multimodal encoders. Another research question of interest is understanding similarities and differences between ULMs (e.g., BERT \cite{bert}) and VLMs \cite{alper2023bert, ChenCDWW23, kamath2023text}. While the recently proposed ULM benchmark, Massive Text Embedding Benchmark (MTEB) \cite{MTEB} highlights the lack of generalization of ULMs on tasks involving text embeddings. They focus solely on text encoders and report an aggregate of different metrics across text-only datasets. Most VLM benchmarks are generated using rule-based algorithms \cite{ma2023crepe, yuksekgonul2023and} and consist of only a pair of sentences (either semantically similar or dissimilar sentences). These similar pairs might not have strong semantic similarities, and the dissimilar pairs can have significant lexical differences, which does not represent a strict setting of evaluation.  Moreover,  we must finetune or linearly probe \cite{liu2023visual} these models to evaluate ULMs or VLMs text encoders using these datasets, which can require significant resources. None of the existing benchmarks systematically evaluates the resilience of model embeddings in the presence of lexical distractors~\cite{iimura2018distractor, taladngoen2022assumptions}, i.e., lexically similar but semantically different negative inputs.

\section{\task Benchmark}

We propose the \task benchmark to evaluate VLMs and ULMs in their ability to understand variance and invariance to semantic and lexical alterations in text. 
%
We leverage datasets \cite{hsieh2023sugarcrepe, liu2023visual} derived from MS-COCO \cite{lin2014microsoft}, a large-scale dataset with images and their corresponding captions, and introduce two novel datasets. 
The process of acquiring the datasets for \task is described in Section \ref{sec:generic-benchmark} and Section \ref{sec:spatial-benchmark}. We further highlight how 
\task
overcomes shortcomings of existing VLM and ULM benchmarks below.

\subsection{Overcoming Shortcomings of Existing Benchmarks}

\textbf{Stricter Evaluation Setting:}
The \task benchmark comprises triplets of sentences
where the first two sentences are semantically equivalent, while the third sentence is semantically opposite but remains lexically close. This setup represents a stricter and more challenging evaluation setting than text pairs that are both semantically and lexically distant. 

\textbf{Direct Comparison of ULMs and VLMs:}
The semantically equivalent text pairs in \task triplets are associated with a semantically-consistent image. This allows us to evaluate VLMs in multimodal and unimodal settings and compare their performance with unimodal text encoders.\\ 

\textbf{Mitigating Ambiguity:}
\task unifies evaluation of VLMs and ULMs to resolve ambiguous findings \citep{alper2023bert, ChenCDWW23} from previous benchmarks, allowing discernment of the weaker encoder among image and text encoders. \\

\textbf{Evaluating Embedding Resilience to Lexical Distractors:}
The \task task contains hard negatives with high lexical overlap with positive captions. This allows us to evaluate the model's resilience to lexical distraction, i.e., its ability to recall positive captions in the presence of lexical distractors present in negative captions. \\

\textbf{Off-the-shelf Evaluation:}
The \task triples and evaluation setting facilitates an off-of-the-shelf evaluation of VLMs and ULMs, unlike previous work \cite{liu2023visual}, that may require further fine-tuning.

In addition, we follow guidelines as mentioned in Appendix \ref{guidelines}, to ensure the applicability of the \task task to both unimodal and VLMs. 

\subsection{Generic \task Benchmark}
\label{sec:generic-benchmark}


To create the generic \task benchmark, we build upon SUGARCREPE benchmark \cite{hsieh2023sugarcrepe}. SUGARCREPE leverages the recent advancements in conditional text generation using LLMs to generate hard negative captions, thereby overcoming the issues with procedurally generated captions. SUGARCREPE consists of (only) one positive and one hard negative caption for each image. The negative captions can range from contradictions in spatial arrangement to contradictions in actions or objects in the image. We expand on their methodology to further introduce an additional positive caption. 
The process of generating the additional positive caption consisted of two parts: 
1) prompting.
2) automated and human validation.


\textbf{Prompting:} We followed an iterative prompting methodology to refine a final prompt that generates optimal second positive captions ($P_2$). Initially, we primed the generative model with a ``role-play" prompt, utilizing the LLama 7b model \cite{touvron2023llama} for our generation process. Prior work, such as \cite{role-playing}, demonstrated that ``role-play" priming can enhance the reasoning abilities of LLMs. We primed the LLM to simulate the role of a ``Data Generating AI", as described in Figure \ref{fig:datagenai-instructions}.
\begin{figure}[tb]
   
    \centering
    \begin{tcolorbox}[
        colback=gray!10, 
        colframe=gray,
        arc=5mm,
        boxrule=1pt,
    ]
      \textbf{Role-playing Prompt:} You are an instruction-following DataGenAI. Your expertise lies in accurately interpreting and generating datasets based on given instructions. Your responses are expected to be precise and limited to the required output.
    \end{tcolorbox}
    \vspace{-7pt}
    \caption{Role playing prompt for ``Data Generator AI".}
    \label{fig:datagenai-instructions}
    \vspace{-10pt}
\end{figure}
The accuracy of LLMs in instruction following can be enhanced by employing explicit and itemized rules, a technique known as ``Rules Prompting" \cite{eval-ins-following}. Following the same strategy, we initially primed the model with the `rules prompt' and subsequently with `demonstrations' that adhere to these rules. The ``rules instruction" and ``demonstrations" are elaborated in Figure \ref{fig:prompting} in Appendix \ref{appendix:generic-data-prompt}.

\textbf{Automated and Human Validation of Generated Captions: }
Using LLMs as evaluators for generated outputs has been extensively explored \cite{llm-judge,eval-ins-following}, providing a cost-effective alternative to human validation. We employed a distinct prompt for automatically verifying semantic consistency between original and generated prompts. The comprehensive prompt for utilizing LLM as an evaluator is detailed in Figure \ref{fig:LLM-evaluation} in the Appendix \ref{app:val_prompt}. The LLM was assigned to evaluate  semantic consistency between positive caption pairs and generate a new caption if inconsistencies were detected. Subsequently, an expert human verified the outputs of the LLM evaluator, selecting the best positive captions among those generated in the prompting and automated evaluation steps, and making minor edits if required. We generated 973 data points comprising triplets (two positives and one negative) associated with images. Figure \ref{fig:generic_ex} in Appendix \ref{appendix:generic_examples} shows examples from the generic \task.


\subsection{Spatial \task Benchmark}
\label{sec:spatial-benchmark}

This dataset is a special case of generic \task benchmark that aims to have two positive captions conveying the same spatial relationships among the objects in the image (see Figure \ref{fig:spatial_ex} in Appendix \ref{spatial_ex} for examples from spatial \task dataset). The negative caption refers to a contradictory spatial arrangement of the objects considered in the positive captions. Below, we detail the dataset creation process that can be divided into three parts: 1) Positive caption ($P_1$, $P_2$) mining; 2) Negative caption ($N$) mining; and 3) Expert Validation and Annotation.

\textbf{Positive Caption Mining: }
To create the spatial \task dataset, we used the Visual Spatial Reasoning (VSR) dataset \cite{liu2023visual}, a subset of COCO  \cite{lin2014microsoft}. In VSR, each image has multiple captions, and the captions are accompanied by descriptive features such as subject, object, and the relationship between them, presented as separate fields. We made use of these descriptive features to select the potential $P_2$ and $N$ captions. 
We utilized subject and object fields to identify the positive pair ($P_1$, $P_2$) for an image. For instance, given an image with caption $P_1$ = \textit{\lq The cat is in the basket\rq}, where $subject=cat$ and $object = basket$, we extracted $P_2$ such that subject and object were switched, resulting in $subject = basket$ and $object = cat$, with caption $P_2$ = \textit{\lq The basket contains the cat\rq}. 

\begin{figure*}[h]
    \centering
    \includegraphics[width=0.91\linewidth]{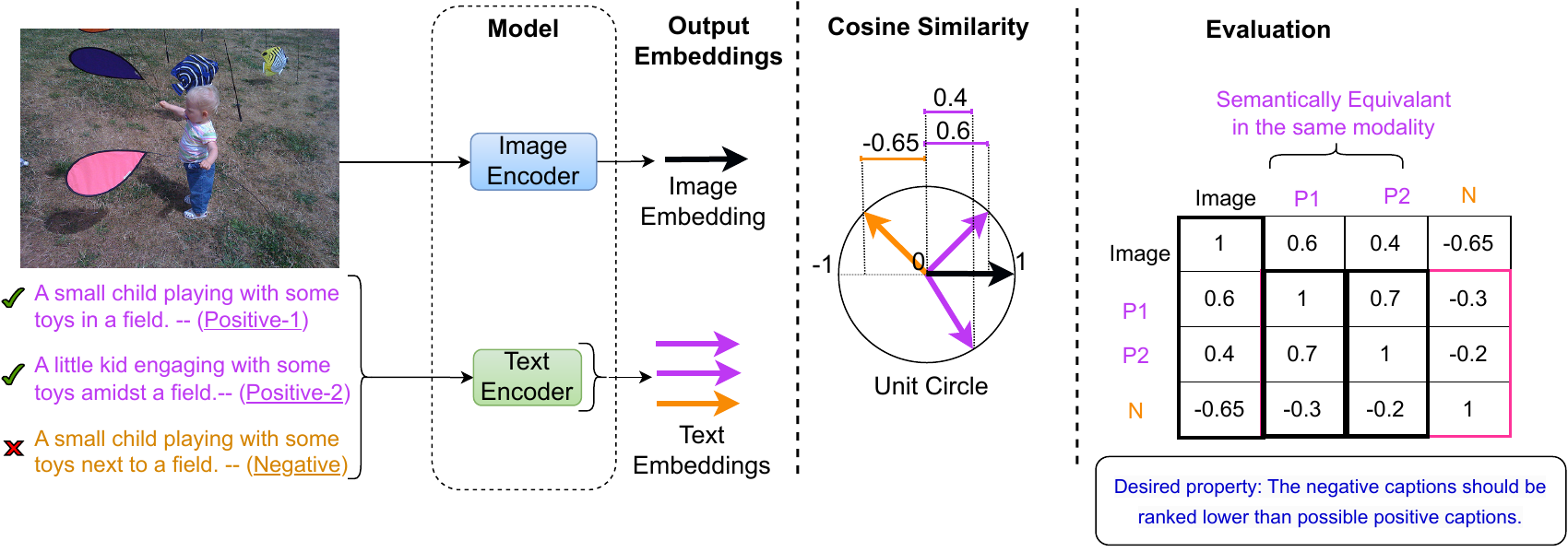}
    \caption{
\small \task task Evaluation: Given an image \(M\) and a triplet of candidate captions $\{$P$_1$, P$_2$, N$\}$ of \(M\), where P$_1$ and P$_2$ are semantically equivalent to each other (referred to as \textit{positive captions} in text), we measure the accuracy of ranking the \textit{negative caption} N below the positive captions for both the Image and Text Encoder.}
    \label{fig:block}
    \vspace{-7pt}
\end{figure*}
\textbf{Negative Caption: }
We extracted an initial negative caption ($N$) by leveraging the observation that a different image containing the same object would likely describe a distinct spatial arrangement between them. Further, we identified the top three negative candidates for each data point by ranking all unique captions in the VSR \cite{liu2023visual} dataset using Jaccard similarity. These provided a set of negative captions candidates that human experts could select or edit.

\textbf{Expert Validation and Annotation: }
To ensure that triplet ($P_1$, $P_2$ and $N$) selected above would adhere to the guidelines defined in Appendix \ref{guidelines}, we performed manual validation and correction of positives and negatives. Final triplets ($P_1$, $P_2$, $N$) were determined by reviewing each data point with input from at least one of four human experts. These experts also ensured that positive pairs ($P_1$, $P_2$) described the same spatial arrangement between objects, and that negatives described the contradictory spatial arrangement. Further, triplets that required image information were marked and modified appropriately. 
Utilizing the described heuristics to generate initial triplets related to the captions significantly reduced the manual effort needed to create triplets from scratch. We note that the majority of data points did require major or minor edits to meet the stringent constraints of a similar spatial arrangement of objects. Finally, we ended  with 640 samples (Image, triplets) satisfying the defined guidelines. Figure \ref{fig:spatial_ex} shows a sample image along with the corresponding triplet from the spatial \task benchmark.
\section{Probing VLMs and ULMs using \task}
\label{exp_res}


We evaluate a comprehensive list of VLMs and ULMs on the \task task that differs in pre-training tasks, pre-training data size, and model size. This includes but is not limited to recent developments in VLMs that aim to improve compositionality of VLMs encoders  \cite{li2022fine, cascante2023going, li2023variational, BugliarelloNH23, ji2023seeing, SinghZWWXDC23} and ULMs \cite{bge_embedding,INSTRUCTOR, wang2023improving, wang2022text} that aims to improve the generalization across tasks involving text embeddings.

\begin{table}[htb]
\caption{\small Different retrieval settings possible with our benchmark and its corresponding positive and negative candidates.}
    \label{tab:retrieval-settings}
    \vspace{0.5cm}
    \centering
    \begin{tabular}{lccc}
        \toprule
        \textbf{Retrieval Setting} & \textbf{Query} & \textbf{Positives} & \textbf{Negatives} \\
        \midrule 
        Text-to-text (T2T) & \{$P_1$/$P_2$\} & \{$P_2$/$P_1$\} & \{$N$\} \\   
        \midrule
        Image-to-text (I2T) & I & \{$P_1$, $P_2$\} & \{$N$\} \\ 
        \bottomrule
    \end{tabular}
    \vspace{-8pt}
\end{table}

\subsection{Experimental Setup}
\label{evaluate}
 We use the text-to-text (T2T) retrieval task to evaluate the text encoders of unimodal models and VLMs on our benchmark. We use the image-to-text (I2T) retrieval task to evaluate the VLMs in a multi-modal setting.
Table \ref{tab:retrieval-settings} illustrates the two retrieval settings for our benchmark, utilizing a tuple of {$I$, $P_1$, $P_2$ and $N$} for the T2T and I2T tasks. The evaluation process involves using triplets ($P_1$, $P_2$ and $N$) to assess text encoders and using the image as a query to evaluate vision encoders. The essential criterion is that the negative caption should be ranked lower than positive captions given either the Image query or Text query. Refer to Figure \ref{fig:block} for an illustration of the evaluation process. We report the accuracy of the model, assigning the last rank to the negative captions for both settings. Our evaluation scheme on \task task using the two settings, is detailed in Appendix \ref{appendix:retrival-settings}.
These evaluations on the \task task provide insights into the performance of VLM and ULM encoders, described below.

\begin{table}[!htb]
    \centering
    \vspace{-5pt}
    \caption{\small Comparison of different ULMs on the Generic and Spatial datasets. We report the Accuracy (\%) of ranking the negative captions last, i.e., below the positive captions. We include the number of parameters in text encoders relative to BERT-base, i.e., 109.5 Million parameters. Overall best values are in bold, and group-level best values are underlined. See Table \ref{tab:unimodal-performance-finegrained_full} in Appendix \ref{appendix:det_ulm} for details.} 
        \label{tab:unimodal-performance}
        \vspace{0.3cm}
    \resizebox{\linewidth}{!}{
        \begin{tabular}{llrr}
            \toprule
            \textbf{Dataset}  &\textbf{\# Params}& \textbf{Generic} &\textbf{Spatial} \\
            \textbf{Model}  &(BERT Scale) & \textbf{Acc(\%)} &\textbf{Acc(\%)} \\
            \toprule
            All-MiniLM-L6-v2  &0.21 &58.89&43.75\\
            Bge-small-en-v1.5 &0.3 &\underline{72.05}&47.97\\
            All-MiniLM-L12-v2  &0.3 &66.19&47.50\\
            GTE-small  &0.3 &64.23&\underline{48.13}\\
            \midrule 
            Angle-BERT-base-uncased-nli-en-v1   &1  &\underline{73.69}&\underline{51.56}\\
            BGE-base-en-v1.5  &1 &72.35&50.16\\
            Sentence-T5-base &1.01  &71.22&\underline{51.09}\\
            GTE-base  &1 &67.42&50.78\\
            Clip-ViT-B-32-multilingual-v1  &1.23 &39.36&38.13\\
            Clip-ViT-B-32  &1.38 &34.64&30.16\\
            \midrule 
            Instructor-large  &3.07 &72.76&\textbf{52.03}\\
            Instructor-large (custom-ins) &3.07 &\underline{75.03}&\textbf{52.81}\\
            UAE-Large-V1  &3.06 &72.97&49.84\\
            GTE-large  &3.06 &68.24&49.06\\
            All-RoBERTa-large-v1   &3.25 &71.94&47.03\\
            Stsb-RoBERTa-large  &3.25 &70.81&50.78\\
            LaBSE  &4.31 &33.92&44.53\\
            Sentence-T5-xl  &11.34 &72.05&51.09\\
            \midrule 
            Angle-Llama-7b-nli-v2 &62.28 &\textbf{78.93}&\textbf{52.34}\\
            E5-Mistral-7b-instruct  &64.95 &\textbf{78.21}&\textbf{52.50}\\
            \bottomrule
        \end{tabular}}
        \vspace{-8pt}
\end{table}

\subsection{Can ULMs understand \tasks?}

There have been recent developments in the quality of text embeddings fueled by progress in language model architectures and training objectives.
We sampled 23 text encoders, covering various model sizes, architectures, and training objectives. For small-sized models, we inspect MiniLM \cite{sbert}, GTE \cite{li2023general} and BGE \cite{bge_embedding}. These are trained on over 1 billion, 200 million, and 803 million sentence pairs, respectively. The results are presented in Table \ref{tab:unimodal-performance}. In this category, BAAI General Embedding (BGE) \cite{bge_embedding} outperforms others in generic \tasks, utilizing an improved contrastive loss objective in a multi-stage fashion. while in spatial \task, GTE-small~\cite{li2023general} is marginally better than BGE, utilizing contrastive learning over a diverse mixture of datasets from multiple sources. 
MiniLM with 12 layers only marginally outperforms MiniLM with 6 layers for generic \tasks, and improves by 3.75\% on spatial \task over the 6-layer variant. In medium-sized models, we observe no improvements in generic \task across all models and a modest increase of +(2.2\%) and +(2.65\%) on the Spatial \task for BGE and GTE, respectively. Suggesting that scale only has a minimal impact on encoding semantics. The BERT model trained with the recently proposed angle objective \cite{li2023angle} outperforms other medium-sized models. Recently proposed to improve generalization across embedding tasks, the Instructor \cite{INSTRUCTOR} model, performs the best among the larger models, albeit with small improvements of 1.34\% and 1.35\% on generic and spatial \tasks respectively. Even the 60 times larger, SOTA generative models like angle-llama-7b-nli-v2 \cite{li2023angle} and e5-mistral-7b-instruct \cite{wang2023improving, wang2022text} especially fine-tuned for emebedding tasks, peak at 52.50\% ACC on the spatial \task and provide only an improvement of +(4.70\%) and +(4.38\%), over the small models in Generic and Spatial \task respectively. These results on the \task tasks indicate the difficulty of ULMs in resolving generic and spatial lexical alteration across all categories, irrespective of architecture, model size, training data size, and optimization objective. Further categorization of errors is presented in the next section.

\begin{table}[tb]
    \centering
    \vspace{-5pt}
    \caption{\small Comparison of ULMs on the Generic and Spatial \task datasets. $P_1$-$N$ and $P_2$-$N$ refer to the accuracy (\%) of ranking positive caption 1 and positive caption 2 above the Negative caption, respectively. $P_1$ have more lexical overlap with $N$. Overall best values are in bold, and group-level best values are underlined.}
        \label{tab:unimodal-performance-finegrained}
        \vspace{0.3cm}
    \resizebox{\linewidth}{!}{
        \begin{tabular}{lllll}
            \toprule
            \textbf{Dataset} &\multicolumn{2}{c}{\textbf{Generic}}& \multicolumn{2}{c}{\textbf{Spatial}}\\ 
            \cmidrule(lr){2-3} \cmidrule(lr){4-5}
            \textbf{Model} &  \textbf{$P_1$-$N$}& \textbf{$P_2$-$N$}& \textbf{$P_1$-$N$}&\textbf{$P_2$-$N$}\\
            \toprule
            All-MiniLM-L6-v2  & 93.22&60.12&\underline{54.38}&46.56\\ 
            Bge-small-en-v1.5 & 93.73&\underline{73.48}&53.28&\underline{52.81}\\ 
            All-MiniLM-L12-v2  & \underline{94.35}&66.91&53.75&50.31\\
            GTE-small  & \underline{94.66}&64.85&\underline{54.69}&\underline{52.19}\\
            \midrule
            Angle-BERT-base-uncased-nli-en-v1   & \underline{94.66}&\underline{74.31}&\underline{55.94}&\underline{58.28}\\ 
            BGE-base-en-v1.5  & \underline{94.14}&73.38&54.38&55.78\\ 
            Sentence-T5-base & \underline{94.24}&72.05&\underline{55}&\underline{58.59}\\ 
            GTE-base  & \underline{94.55}&67.73&\underline{55.63}&54.22\\ 
            Clip-ViT-B-32-multilingual-v1  & 78.83&40.90&52.97&45.47\\ 
            Clip-ViT-B-32  & 81.71&35.46&52.66&35.78\\
            \midrule
            Instructor-large  & 93.63&73.48&\textbf{56.72}&57.66\\
 Instructor-large (custom-ins) & \underline{94.86}& \underline{75.64}& \textbf{56.09}&61.25\\ 
            UAE-Large-V1  & \underline{94.14}&74.20&54.69&58.59\\ 
            GTE-large  & \underline{94.76}&68.65&55.16&57.66\\  
            All-RoBERTa-large-v1   & 93.94&72.97&55&52.5\\ 
            Stsb-RoBERTa-large  & 93.01&71.63&54.22&\textbf{66.88}\\ 
            LaBSE  & 84.89&35.05&53.28&45.94\\ 
            Sentence-T5-xl  & \underline{94.55}&72.66&55.31&63.75\\
            \midrule
            Angle-Llama-7b-nli-v2 & \textbf{95.89}&\textbf{79.34}&\textbf{56.41}&\underline{61.41}\\
 E5-Mistral-7b-instruct  & \textbf{95.58}& 78.93& 55.31&60.16\\
            \bottomrule
        \end{tabular}}
        \vspace{-10pt}
    \end{table}

    \begin{table}[htb]
\caption{\small Comparison of VLMs performance when tested on the generic and spatial paraphrasing benchmarks. Performance reported in terms of Accuracy (\%). VLM: Both vision and text encoder embeddings compared; Text: only text encoder embeddings compared. XVLM-ITR-COCO and XVLM-ITR-Flickr are finetuned on XVLM-16M models. T2T and I2T refer to text-to-text and image-to-text retrieval, respectively. Overall best values are in bold, and group-level best values are underlined.}
\label{tab:gen_spat}
\footnotesize
\centering
\vspace{0.2cm}
\begin{tabular}{l|cccc}
\toprule
& \multicolumn{2}{c}{Generic} & \multicolumn{2}{c}{Spatial} \\
\cmidrule(lr){2-3} \cmidrule(lr){4-5}
Model &  T2T & I2T & T2T & I2T \\
\midrule 
CLIP-ViT-B/32 & 34.63 & 54.88 & 30.16 & \underline{44.69} \\
RoBERTa-ViT-B/32 & \underline{53.34} & \underline{58.89} & \underline{36.25} & 37.66 \\
ALIGN & 42.34 & 50.46 & 34.53 & 35.16 \\
ALIP & 21.07 & 47.79 & 17.82 & 38.75 \\
\midrule
FLAVA & \underline{56.32} & \underline{59.81} & 28.44 & 25.31 \\
ALBEF & 28.26 & 49.54 & 25.78 & 42.66 \\
BLIP & 44.89 & 54.21 & 39.38 & \underline{45.63} \\
BLIP2 & 41.11 & 51.69 & \underline{40.62} & 41.09 \\
\midrule
ViLT & -- & 33.61 & -- & 20.32 \\
AltCLIP & \underline{53.44} & \underline{57.76} & \underline{35.16} & \underline{45.01} \\
\midrule
SegCLIP & 37.54 & 57.07 & 25.63 & 33.59 \\
XVLM-4M & 28.57 & 47.69 & 24.84 & 42.19 \\
XVLM-16M & \underline{47.58} & \underline{59.40} & \underline{31.41} & \underline{50.31} \\
\midrule
BLIP-ITM-COCO & -- & 58.38 & -- & 33.59 \\
ViLT-ITR-COCO & -- & 61.04 & -- & 50.16 \\
XVLM-ITR-COCO & \textbf{61.56} & \textbf{62.38} & \textbf{45.16} & \textbf{51.09} \\
XVLM-ITR-Flickr & 58.68 & 62.07 & 39.69 & 45.16 \\
\midrule
NegCLIP & \underline{48.72} & \underline{52.83} & 29.21 & \underline{34.84} \\
CLIP-SVLC & 48.11 & 44.60 & \underline{30.94} & 28.75 \\
CyCLIP & 26.31 & 38.23 & 12.50 & 31.41 \\
BLIP-SGVL & -- & 28.88 & -- & 33.75 \\
\bottomrule
\end{tabular}
\vspace{-8pt}
\end{table}


\subsection{Can lexical distractors overpower semantics?}

We know positive captions exhibit substantial semantic alignment, while negative captions show minimal semantic alignment with their positive counterparts. In retrieval, a rudimentary scenario arises when the query and negative candidates differ both in semantics and lexicon. Our dataset is designed such that positive and negative instances deliberately showcase significant lexical overlap, while the positives show minimal lexical overlap between them. Both of these cases capture two challenging retrieval aspects in terms of semantics. We illustrate this phenomenon in Figure \ref{fig:dist_comb_labs} in Appendix \ref{appendix:lexical_overlap}, using edit distance as a proxy for lexical overlap. For a given triplet, we reorder the positive captions so that the first positive caption, i.e., P1, exhibits a higher lexical overlap with the negative caption (N). This provides a controlled setting to inspect whether embeddings are more sensitive to lexical or semantic changes. Table \ref{tab:unimodal-performance-finegrained} shows the results of this analysis. A notable observation is the evident separation in the model performances of P1-N and P2-N accuracies, i.e.,  the accuracy of the first caption ranking higher than the negative caption and the accuracy of the second positive caption ranking higher than the negative caption. Interestingly. We observe that P1-N accuracy is always higher than P2-N accuracy for the generic \task  and not always higher than P2-N for the spatial \task. This suggests that spatial understanding in language models is highly sensitive to lexical information, and lexical overlap can divert models from capturing spatial semantics. In contrast, lexical overlap among the candidates does not hinder the correct recall of semantically equivalent options of a text query. Implying not all kinds of semantics are treated equally by the embeddings, especially spatial semantics.

\subsection{How visual information effects \task in VLMs?}

\textbf{Analyzed models:} We comprehensively evaluate a wide array of VLMs, which include: 1) Models trained with a contrastive learning objective such as CLIP-ViT-B/32~\cite{radford2021learning}, RoBERTa-ViT-B/32~\cite{schuhmann2022laion}, ALIGN~\cite{jia2021scaling} and ALIP~\cite{yang2023alip}. 
2) Models trained by combining multiple objective functions, such as FLAVA~\cite{singh2022flava},  ALBEF~\cite{li2021albef}, BLIP~\cite{li2022blip} and BLIP-2~\cite{li2023blip2}.
3) Models with a unified encoder for text and images, such as ViLT~\cite{kim2021vilt}, and multi-lingual distilled models like AltCLIP~\cite{ChenLZYW23}; 4) Models that align text with corresponding visual concepts in the image, such as SegCLIP~\cite{luo2023segclip}, and XVLM~\cite{zeng2022multi} - with two variants, XVLM-4M and XVLM-16M.

We also investigate several models that have been finetuned on downstream tasks of image-text retrieval, such as BLIP-ITM-COCO~\cite{li2022blip}, ViLT-ITR-COCO \cite{kim2021vilt} and XVLM-16M-ITR-COCO \cite{zeng2022multi}. Specifically, BLIP, ViLT, and XVLM-16M models were trained for the ITM task using the COCO dataset. Additionally, XVLM-16M-ITR-Flickr \cite{zeng2022multi} denotes XVLM-16M models trained for the ITM task using the Flickr dataset. Moreover, we evaluate recent methods proposed to improve the compositionality of VLMs, including NegCLIP \cite{yuksekgonul2023and}, SVLC \cite{doveh2023teaching}, CyCLIP \cite{goel2022cyclip}, and BLIP-SGVL \cite{HerzigMKAFDG23}. 
These models differ in terms of model architecture, total number of parameters and embedding dimension and pretraining objectives and more details are in Table \ref{tab:vlm_details} of Appendix \ref{appendix:vlm_eval}.

\textbf{Results:} We evaluate the VLMs using the \task datasets in two distinct ways, T2T and I2T, as explained in Section \ref{evaluate}.
Table \ref{tab:gen_spat} provides a comparison between different VLMs. (See Tables \ref{tab:generic} and \ref{tab:spatial}in Appendix \ref{det_res_vlm} for detailed results)

\underline{T2T task}: Among the models pretrained using contrastive loss, RoBERTa-Vit-B/32 performs better on both benchmarks. Models trained with multiple objective functions show better performance than those trained solely with a contrastive loss function. In particular, BLIP and BLIP-2 models, pretrained with contrastive, ITM, and image captioning objectives, achieve superior performance on both benchmarks. Additionally, text encoders of the models pretrained to align text with corresponding visual concepts in images perform better than CLIP-based text encoders. This indicates that the contrastive learning objective alone may not be sufficient for text encoders to learn the semantic relations between text and image. Interestingly, models proposed to improve compositionality, such as NegCLIP and CLIP-SVLC, achieve better performance than CLIP, underscoring the importance of compositionality for the \task task. Models fine-tuned on the downstream task of ITR, particularly on the COCO dataset, achieve the best performance.
Furthermore, for all VLMs, a significant drop in performance is observed for the spatial benchmark compared to the generic benchmark, suggesting that the text encoders of VLMs struggle with understanding spatial VISLA.

When compared with unimodal text encoders (see Table \ref{tab:unimodal-performance}), the performance of all the VLMs falls behind on the \task task. This indicates that the text encoders of VLMs are more sensitive to the semantic and lexical variations compared to the unimodal text encoders.  

\underline{I2T task}: All VLMs achieve higher performance on the I2T task compared to the T2T task on both benchmarks. This demonstrates that the inclusion of visual information improves the performance of VLMs on the \task task. Similar to T2T task, models trained using multiple objective functions perform better than models trained using only contrastive loss on I2T task. Both models trained exclusively with contrastive loss and those trained using multiple objective functions show significant improvements in performance on the I2T task compared to the T2T task. The model pretrained to align text with semantic concepts of images performs better than CLIP, indicating that better semantic understanding is beneficial for the \task task. 
However, models trained to improve compositionality do not show consistent improvements in performance on the I2T task compared to the T2T task. Moreover, the performance of these models is comparable or lower than CLIP models. Particularly on the spatial benchmark, these models show a degradation in performance compared to CLIP.

\subsection{Does model and data size effect \task?}

To address this question, we investigated various CLIP~\cite{radford2021learning} variants trained on the WebImageText dataset, which comprises 400 million image-text pairs. These models encompass CNN-based architectures, such as RN50, RN101, RN$50\times4$, RN$50\times16$, and RN$50\times64$, as well as transformer-based models like ViT-B/32 and ViT-L/14. Additionally, we examined CLIP-based models introduced by \cite{schuhmann2022laion} and \cite{gadre2023datacomp}, pre-trained on extensive paired image-text datasets. \citet{schuhmann2022laion} provided diverse CLIP variants, namely RoBERTa-ViT-B/32, ViT-H/14, ViT-g/14, xlm-roberta-base-ViT-B/32, and xlm-roberta-large-ViT-H/14, trained on a large image-text dataset called "LAION-5B," which consists of 5 billion image-text pairs. Similarly, \citet{gadre2023datacomp} released two CLIP variants, namely Large:ViT-B/16 and xlarge:ViT-L/14, trained on the DataComp dataset, comprising 13 billion image-text pairs.

The performance of various CLIP variants on \task benchmarks is detailed in Table \ref{tab:clip_variants} (in Appendix \ref{app:clip_var}). For both I2T and T2T tasks, we observed no significant variations in performance among the different CLIP variants trained on a dataset of 400 million image-text pairs. In contrast, CLIP variants trained using the LAION dataset, comprising 2 billion image-text pairs, demonstrated superior performance on the Generic benchmark compared to those trained on the 400 million-sample dataset. However, increasing the LAION training data size from 2 billion to 5 billion did not result in performance improvement. Similar trends were noted in models trained on the DataComp dataset. Notably, all CLIP variants, regardless of the training data size, exhibited lower performance on the spatial benchmark compared to the Generic benchmark. 

\subsection{Understanding difficulties of VLMs on VISLA}
To understand the low performance of VLMs on \task, we analyzed the semantically equivalent and semantically different pairs in the \task benchmark that confuse VLMs under the two evaluation settings -- I2T and T2T.


Figure \ref{fig:err_img_txt_vlm_sc} (in Appendix \ref{qual_vlm_1}) highlights examples for which the VLMs (I2T task: both image and text as input) failed in identifying the correct captions ($P_1$, $P_2$) given the image ($I$). Specifically, the VLMs place the negative caption ($N$) closer to the image($I$) than both the positive captions. Lexical alterations such as swapping the subject and the object and replacing words with their synonyms/antonyms can mislead the VLMs when tested in the multimodal setting. 


Figure \ref{fig:img_cor_txt_error_vlm} (in Appendix \ref{qual_vlm_2}) provides the examples for which the VLMs succeed on the I2T task but fail to identify the semantically equivalent pairs when exclusively evaluating the text encoder of the VLMs. We observe that even lexical alterations, such as simple reordering of words using synonyms and antonyms, can fool the VLM text encoders from identifying the semantically equivalent pairs from the \task triples. These observations may suggest that the text encoders of VLMs struggle to differentiate semantics from syntactics. This is consistent with the observation in \cite{kamath2023text}

\section{Conclusion}
In this work, we evaluate a comprehensive list of VLMs and unimodal language models on the Variance and Invariance to Semantic and Lexical Alterations (VISLA) task by introducing two new datasets. We show that unimodal text encoders have difficulties with \task task in general.
Unimodal text encoders perform moderately on generic \task benchmark but show significant degradation in performance on spatial \task benchmark. 

We also show that the text encoders of VLMs perform inferior to ULM text encoders. The performance of the VLMs on \task task is better under the multimodal setting when compared to the unimodal text-only setting. Interestingly, varying the model architecture or size of the VLMs will not improve the performance on the \task benchmark. On the other hand, increasing the pre-training dataset size is shown to improve the performance of the VLMs on VISLA. Further qualitative analysis shows that the text encoders of VLMs struggle to understand variance and invariance of semantics to simple lexical alterations.

\section*{Impact Statement}
This paper presents work whose goal is to advance the field of Machine Learning in general and Language Models research in particular. We discuss several limitations of language models related to the separation of semantics of an input text from its syntactic and lexical form. In order to build trust-worthy Language Models, it is important to establish that the language models emphasize semantics contained in a sentence rather than the lexical form and syntactic style of the sentence. Our evaluation provides evidence of this problem through two curated datasets and can potentially be impactful for evaluating newer language models and/or inspiring novel solutions to this problem. There are many other potential societal consequences of our work, none of which we feel must be specifically highlighted here.

\section*{Acknowledgements}
We thank the Canadian Institute for Advanced Research (CIFAR) for their support. Resources used in preparing this research were provided, in part, by NSERC, the Province of Ontario, the Government of Canada through CIFAR, and companies sponsoring the Vector Institute \url{www.vectorinstitute.ai/#partners}.

\bibliographystyle{abbrvnat}
\bibliography{references}

\begin{thebibliography}{78}
\providecommand{\natexlab}[1]{#1}
\providecommand{\url}[1]{\texttt{#1}}
\expandafter\ifx\csname urlstyle\endcsname\relax
  \providecommand{\doi}[1]{doi: #1}\else
  \providecommand{\doi}{doi: \begingroup \urlstyle{rm}\Url}\fi

\bibitem[Agirre et~al.(2012)Agirre, Diab, Cer, and Gonzalez-Agirre]{sts12t6}
E.~Agirre, M.~Diab, D.~Cer, and A.~Gonzalez-Agirre.
\newblock Semeval-2012 task 6: a pilot on semantic textual similarity.
\newblock In \emph{Proceedings of the First Joint Conference on Lexical and Computational Semantics - Volume 1: Proceedings of the Main Conference and the Shared Task, and Volume 2: Proceedings of the Sixth International Workshop on Semantic Evaluation}, SemEval '12, page 385–393, USA, 2012. Association for Computational Linguistics.

\bibitem[Agirre et~al.(2013)Agirre, Cer, Diab, Gonzalez-Agirre, and Guo]{sts13}
E.~Agirre, D.~Cer, M.~Diab, A.~Gonzalez-Agirre, and W.~Guo.
\newblock *{SEM} 2013 shared task: Semantic textual similarity.
\newblock In M.~Diab, T.~Baldwin, and M.~Baroni, editors, \emph{Second Joint Conference on Lexical and Computational Semantics (*{SEM}), Volume 1: Proceedings of the Main Conference and the Shared Task: Semantic Textual Similarity}, pages 32--43, Atlanta, Georgia, USA, June 2013. Association for Computational Linguistics.
\newblock URL \url{https://aclanthology.org/S13-1004}.

\bibitem[Agirre et~al.(2014)Agirre, Banea, Cardie, Cer, Diab, Gonzalez-Agirre, Guo, Mihalcea, Rigau, and Wiebe]{sts14t10}
E.~Agirre, C.~Banea, C.~Cardie, D.~Cer, M.~Diab, A.~Gonzalez-Agirre, W.~Guo, R.~Mihalcea, G.~Rigau, and J.~Wiebe.
\newblock {S}em{E}val-2014 task 10: Multilingual semantic textual similarity.
\newblock In P.~Nakov and T.~Zesch, editors, \emph{Proceedings of the 8th International Workshop on Semantic Evaluation ({S}em{E}val 2014)}, pages 81--91, Dublin, Ireland, Aug. 2014. Association for Computational Linguistics.
\newblock \doi{10.3115/v1/S14-2010}.
\newblock URL \url{https://aclanthology.org/S14-2010}.

\bibitem[Agirre et~al.(2015)Agirre, Banea, Cardie, Cer, Diab, Gonzalez-Agirre, Guo, Lopez-Gazpio, Maritxalar, Mihalcea, Rigau, Uria, and Wiebe]{sts15t2}
E.~Agirre, C.~Banea, C.~Cardie, D.~Cer, M.~Diab, A.~Gonzalez-Agirre, W.~Guo, I.~Lopez-Gazpio, M.~Maritxalar, R.~Mihalcea, G.~Rigau, L.~Uria, and J.~Wiebe.
\newblock {S}em{E}val-2015 task 2: Semantic textual similarity, {E}nglish, {S}panish and pilot on interpretability.
\newblock In P.~Nakov, T.~Zesch, D.~Cer, and D.~Jurgens, editors, \emph{Proceedings of the 9th International Workshop on Semantic Evaluation ({S}em{E}val 2015)}, pages 252--263, Denver, Colorado, June 2015. Association for Computational Linguistics.
\newblock \doi{10.18653/v1/S15-2045}.
\newblock URL \url{https://aclanthology.org/S15-2045}.

\bibitem[Agirre et~al.(2016)Agirre, Banea, Cer, Diab, Gonzalez-Agirre, Mihalcea, Rigau, and Wiebe]{sts16t1}
E.~Agirre, C.~Banea, D.~Cer, M.~Diab, A.~Gonzalez-Agirre, R.~Mihalcea, G.~Rigau, and J.~Wiebe.
\newblock {S}em{E}val-2016 task 1: Semantic textual similarity, monolingual and cross-lingual evaluation.
\newblock In S.~Bethard, M.~Carpuat, D.~Cer, D.~Jurgens, P.~Nakov, and T.~Zesch, editors, \emph{Proceedings of the 10th International Workshop on Semantic Evaluation ({S}em{E}val-2016)}, pages 497--511, San Diego, California, June 2016. Association for Computational Linguistics.
\newblock \doi{10.18653/v1/S16-1081}.
\newblock URL \url{https://aclanthology.org/S16-1081}.

\bibitem[Alper et~al.(2023)Alper, Fiman, and Averbuch-Elor]{alper2023bert}
M.~Alper, M.~Fiman, and H.~Averbuch-Elor.
\newblock Is bert blind? exploring the effect of vision-and-language pretraining on visual language understanding.
\newblock In \emph{Proceedings of the IEEE/CVF Conference on Computer Vision and Pattern Recognition}, pages 6778--6788, 2023.

\bibitem[Bugliarello et~al.(2023)Bugliarello, Nematzadeh, and Hendricks]{BugliarelloNH23}
E.~Bugliarello, A.~Nematzadeh, and L.~A. Hendricks.
\newblock Weakly-supervised learning of visual relations in multimodal pretraining.
\newblock In \emph{Proceedings of the 2023 Conference on Empirical Methods in Natural Language Processing, {EMNLP} 2023}, pages 3052--3071. Association for Computational Linguistics, 2023.

\bibitem[Cao et~al.(2021)Cao, Aziz, and Titov]{p0}
N.~D. Cao, W.~Aziz, and I.~Titov.
\newblock Editing factual knowledge in language models.
\newblock In M.~Moens, X.~Huang, L.~Specia, and S.~W. Yih, editors, \emph{Proceedings of the 2021 Conference on Empirical Methods in Natural Language Processing, {EMNLP} 2021, Virtual Event / Punta Cana, Dominican Republic, 7-11 November, 2021}, pages 6491--6506. Association for Computational Linguistics, 2021.
\newblock \doi{10.18653/V1/2021.EMNLP-MAIN.522}.
\newblock URL \url{https://doi.org/10.18653/v1/2021.emnlp-main.522}.

\bibitem[Cascante-Bonilla et~al.(2023)Cascante-Bonilla, Shehada, Smith, Doveh, Kim, Panda, Varol, Oliva, Ordonez, Feris, et~al.]{cascante2023going}
P.~Cascante-Bonilla, K.~Shehada, J.~S. Smith, S.~Doveh, D.~Kim, R.~Panda, G.~Varol, A.~Oliva, V.~Ordonez, R.~Feris, et~al.
\newblock Going beyond nouns with vision \& language models using synthetic data.
\newblock In \emph{Proceedings of the IEEE/CVF International Conference on Computer Vision}, pages 20155--20165, 2023.

\bibitem[Cer et~al.(2017)Cer, Diab, Agirre, Lopez-Gazpio, and Specia]{sts17t1}
D.~Cer, M.~Diab, E.~Agirre, I.~Lopez-Gazpio, and L.~Specia.
\newblock {S}em{E}val-2017 task 1: Semantic textual similarity multilingual and crosslingual focused evaluation.
\newblock In S.~Bethard, M.~Carpuat, M.~Apidianaki, S.~M. Mohammad, D.~Cer, and D.~Jurgens, editors, \emph{Proceedings of the 11th International Workshop on Semantic Evaluation ({S}em{E}val-2017)}, pages 1--14, Vancouver, Canada, Aug. 2017. Association for Computational Linguistics.

\bibitem[Chen et~al.(2023{\natexlab{a}})Chen, Chen, Diao, Wan, and Wang]{ChenCDWW23}
Z.~Chen, G.~Chen, S.~Diao, X.~Wan, and B.~Wang.
\newblock On the difference of bert-style and clip-style text encoders.
\newblock In \emph{Findings of the Association for Computational Linguistics: {ACL} 2023}, pages 13710--13721, 2023{\natexlab{a}}.

\bibitem[Chen et~al.(2023{\natexlab{b}})Chen, Liu, Zhang, Yang, and Wu]{ChenLZYW23}
Z.~Chen, G.~Liu, B.~Zhang, Q.~Yang, and L.~Wu.
\newblock Altclip: Altering the language encoder in {CLIP} for extended language capabilities.
\newblock In \emph{Findings of the Association for Computational Linguistics: {ACL} 2023}, pages 8666--8682. Association for Computational Linguistics, 2023{\natexlab{b}}.

\bibitem[Chow et~al.(2023)Chow, Tan, and Kan]{chow2023travlr}
K.~J. Chow, S.~Tan, and M.-Y. Kan.
\newblock Travlr: Now you see it, now you don’t! a bimodal dataset for evaluating visio-linguistic reasoning.
\newblock In \emph{Proceedings of the 17th Conference of the European Chapter of the Association for Computational Linguistics}, pages 3314--3339, 2023.

\bibitem[Devlin et~al.(2019)Devlin, Chang, Lee, and Toutanova]{bert}
J.~Devlin, M.~Chang, K.~Lee, and K.~Toutanova.
\newblock {BERT:} pre-training of deep bidirectional transformers for language understanding.
\newblock In \emph{Proceedings of the 2019 Conference of the North American Chapter of the Association for Computational Linguistics: Human Language Technologies, {NAACL-HLT} 2019}, pages 4171--4186. Association for Computational Linguistics, 2019.

\bibitem[Ding et~al.(2022)Ding, Xue, Xia, and Dai]{Ding0XD22}
J.~Ding, N.~Xue, G.~Xia, and D.~Dai.
\newblock Decoupling zero-shot semantic segmentation.
\newblock In \emph{{IEEE/CVF} Conference on Computer Vision and Pattern Recognition, {CVPR} 2022}, pages 11573--11582. {IEEE}, 2022.

\bibitem[Diwan et~al.(2022)Diwan, Berry, Choi, Harwath, and Mahowald]{DiwanBCHM22}
A.~Diwan, L.~Berry, E.~Choi, D.~Harwath, and K.~Mahowald.
\newblock Why is winoground hard? investigating failures in visuolinguistic compositionality.
\newblock In \emph{Proceedings of the 2022 Conference on Empirical Methods in Natural Language Processing, {EMNLP} 2022}, pages 2236--2250. Association for Computational Linguistics, 2022.

\bibitem[Dolan and Brockett(2005)]{dolan2005automatically}
B.~Dolan and C.~Brockett.
\newblock Automatically constructing a corpus of sentential paraphrases.
\newblock In \emph{Third International Workshop on Paraphrasing (IWP2005)}, 2005.

\bibitem[Doveh et~al.(2023)Doveh, Arbelle, Harary, Schwartz, Herzig, Giryes, Feris, Panda, Ullman, and Karlinsky]{doveh2023teaching}
S.~Doveh, A.~Arbelle, S.~Harary, E.~Schwartz, R.~Herzig, R.~Giryes, R.~Feris, R.~Panda, S.~Ullman, and L.~Karlinsky.
\newblock Teaching structured vision \& language concepts to vision \& language models.
\newblock In \emph{Proceedings of the IEEE/CVF Conference on Computer Vision and Pattern Recognition}, pages 2657--2668, 2023.

\bibitem[Feng et~al.(2022)Feng, Yang, Cer, Arivazhagan, and Wang]{labse}
F.~Feng, Y.~Yang, D.~Cer, N.~Arivazhagan, and W.~Wang.
\newblock Language-agnostic {BERT} sentence embedding.
\newblock In \emph{Proceedings of the 60th Annual Meeting of the Association for Computational Linguistics (Volume 1: Long Papers), {ACL} 2022}, pages 878--891. Association for Computational Linguistics, 2022.

\bibitem[Gadre et~al.(2023)Gadre, Ilharco, Fang, Hayase, Smyrnis, Nguyen, Marten, Wortsman, Ghosh, Zhang, et~al.]{gadre2023datacomp}
S.~Y. Gadre, G.~Ilharco, A.~Fang, J.~Hayase, G.~Smyrnis, T.~Nguyen, R.~Marten, M.~Wortsman, D.~Ghosh, J.~Zhang, et~al.
\newblock Datacomp: In search of the next generation of multimodal datasets.
\newblock \emph{arXiv preprint arXiv:2304.14108}, 2023.

\bibitem[Goel et~al.(2022)Goel, Bansal, Bhatia, Rossi, Vinay, and Grover]{goel2022cyclip}
S.~Goel, H.~Bansal, S.~Bhatia, R.~Rossi, V.~Vinay, and A.~Grover.
\newblock Cyclip: Cyclic contrastive language-image pretraining.
\newblock \emph{Advances in Neural Information Processing Systems}, 35:\penalty0 6704--6719, 2022.

\bibitem[Herzig et~al.(2023)Herzig, Mendelson, Karlinsky, Arbelle, Feris, Darrell, and Globerson]{HerzigMKAFDG23}
R.~Herzig, A.~Mendelson, L.~Karlinsky, A.~Arbelle, R.~Feris, T.~Darrell, and A.~Globerson.
\newblock Incorporating structured representations into pretrained vision {\&} language models using scene graphs.
\newblock In \emph{Proceedings of the 2023 Conference on Empirical Methods in Natural Language Processing, {EMNLP} 2023}, pages 14077--14098. Association for Computational Linguistics, 2023.

\bibitem[Hsieh et~al.(2023)Hsieh, Zhang, Ma, Kembhavi, and Krishna]{hsieh2023sugarcrepe}
C.-Y. Hsieh, J.~Zhang, Z.~Ma, A.~Kembhavi, and R.~Krishna.
\newblock Sugarcrepe: Fixing hackable benchmarks for vision-language compositionality.
\newblock In \emph{Thirty-Seventh Conference on Neural Information Processing Systems Datasets and Benchmarks Track}, 2023.

\bibitem[IIMURA(2018)]{iimura2018distractor}
H.~IIMURA.
\newblock Distractor plausibility in a multiple-choice listening test.
\newblock \emph{JLTA Journal}, 21:\penalty0 65--81, 2018.

\bibitem[Iyer et~al.(2017)Iyer, Dandekar, and Csernai]{qqp_link}
S.~Iyer, N.~Dandekar, and K.~Csernai.
\newblock First quora dataset release: Question pairs, 2017.
\newblock URL \url{https://quoradata.quora.com/First-Quora-Dataset-Release-Question-Pairs}.
\newblock Accessed: 2024-01-01.

\bibitem[Ji et~al.(2023)Ji, Tu, Jiang, Kong, Cai, Zhao, Wang, Yang, and Liu]{ji2023seeing}
Y.~Ji, R.~Tu, J.~Jiang, W.~Kong, C.~Cai, W.~Zhao, H.~Wang, Y.~Yang, and W.~Liu.
\newblock Seeing what you miss: Vision-language pre-training with semantic completion learning.
\newblock In \emph{Proceedings of the IEEE/CVF Conference on Computer Vision and Pattern Recognition}, pages 6789--6798, 2023.

\bibitem[Jia et~al.(2021)Jia, Yang, Xia, Chen, Parekh, Pham, Le, Sung, Li, and Duerig]{jia2021scaling}
C.~Jia, Y.~Yang, Y.~Xia, Y.-T. Chen, Z.~Parekh, H.~Pham, Q.~Le, Y.-H. Sung, Z.~Li, and T.~Duerig.
\newblock Scaling up visual and vision-language representation learning with noisy text supervision.
\newblock In \emph{International conference on machine learning}, pages 4904--4916. PMLR, 2021.

\bibitem[Kamath et~al.(2023)Kamath, Hessel, and Chang]{kamath2023text}
A.~Kamath, J.~Hessel, and K.-W. Chang.
\newblock Text encoders bottleneck compositionality in contrastive vision-language models.
\newblock In \emph{Proceedings of the 2023 Conference on Empirical Methods in Natural Language Processing}, pages 4933--4944, 2023.

\bibitem[Kim et~al.(2021)Kim, Son, and Kim]{kim2021vilt}
W.~Kim, B.~Son, and I.~Kim.
\newblock Vilt: Vision-and-language transformer without convolution or region supervision.
\newblock In \emph{International Conference on Machine Learning}, pages 5583--5594. PMLR, 2021.

\bibitem[Kong et~al.(2023)Kong, Zhao, Chen, Li, Qin, Sun, and Zhou]{role-playing}
A.~Kong, S.~Zhao, H.~Chen, Q.~Li, Y.~Qin, R.~Sun, and X.~Zhou.
\newblock Better zero-shot reasoning with role-play prompting.
\newblock \emph{CoRR}, abs/2308.07702, 2023.

\bibitem[Krishna et~al.(2023)Krishna, Song, Karpinska, Wieting, and Iyyer]{krishna2023paraphrasing}
K.~Krishna, Y.~Song, M.~Karpinska, J.~F. Wieting, and M.~Iyyer.
\newblock Paraphrasing evades detectors of {AI}-generated text, but retrieval is an effective defense.
\newblock In \emph{Thirty-seventh Conference on Neural Information Processing Systems}, 2023.
\newblock URL \url{https://openreview.net/forum?id=WbFhFvjjKj}.

\bibitem[Li et~al.(2021)Li, Selvaraju, Gotmare, Joty, Xiong, and Hoi]{li2021albef}
J.~Li, R.~Selvaraju, A.~Gotmare, S.~Joty, C.~Xiong, and S.~C.~H. Hoi.
\newblock Align before fuse: Vision and language representation learning with momentum distillation.
\newblock \emph{Advances in neural information processing systems}, 34:\penalty0 9694--9705, 2021.

\bibitem[Li et~al.(2022{\natexlab{a}})Li, He, Wei, Qian, Zhu, Xie, Zhuang, Tian, and Tang]{li2022fine}
J.~Li, X.~He, L.~Wei, L.~Qian, L.~Zhu, L.~Xie, Y.~Zhuang, Q.~Tian, and S.~Tang.
\newblock Fine-grained semantically aligned vision-language pre-training.
\newblock \emph{Advances in neural information processing systems}, 35:\penalty0 7290--7303, 2022{\natexlab{a}}.

\bibitem[Li et~al.(2022{\natexlab{b}})Li, Li, Xiong, and Hoi]{li2022blip}
J.~Li, D.~Li, C.~Xiong, and S.~Hoi.
\newblock Blip: Bootstrapping language-image pre-training for unified vision-language understanding and generation.
\newblock In \emph{International Conference on Machine Learning}, pages 12888--12900. PMLR, 2022{\natexlab{b}}.

\bibitem[Li et~al.(2023{\natexlab{a}})Li, Li, Savarese, and Hoi]{li2023blip2}
J.~Li, D.~Li, S.~Savarese, and S.~Hoi.
\newblock {BLIP-2:} bootstrapping language-image pre-training with frozen image encoders and large language models.
\newblock In \emph{ICML}, 2023{\natexlab{a}}.

\bibitem[Li et~al.(2023{\natexlab{b}})Li, Tang, Zhu, Zhang, Yang, Chua, and Wu]{li2023variational}
J.~Li, S.~Tang, L.~Zhu, W.~Zhang, Y.~Yang, T.-S. Chua, and F.~Wu.
\newblock Variational cross-graph reasoning and adaptive structured semantics learning for compositional temporal grounding.
\newblock \emph{IEEE Transactions on Pattern Analysis and Machine Intelligence}, 2023{\natexlab{b}}.

\bibitem[Li and Li(2023)]{li2023angle}
X.~Li and J.~Li.
\newblock Angle-optimized text embeddings.
\newblock \emph{arXiv preprint arXiv:2309.12871}, 2023.

\bibitem[Li et~al.(2023{\natexlab{c}})Li, Zhang, Zhang, Long, Xie, and Zhang]{li2023general}
Z.~Li, X.~Zhang, Y.~Zhang, D.~Long, P.~Xie, and M.~Zhang.
\newblock Towards general text embeddings with multi-stage contrastive learning, 2023{\natexlab{c}}.

\bibitem[Liang et~al.(2023)Liang, Wu, Dai, Li, Zhao, Zhang, Zhang, Vajda, and Marculescu]{LiangWDLZ0ZVM23}
F.~Liang, B.~Wu, X.~Dai, K.~Li, Y.~Zhao, H.~Zhang, P.~Zhang, P.~Vajda, and D.~Marculescu.
\newblock Open-vocabulary semantic segmentation with mask-adapted {CLIP}.
\newblock In \emph{{IEEE/CVF} Conference on Computer Vision and Pattern Recognition, {CVPR} 2023}, pages 7061--7070. {IEEE}, 2023.

\bibitem[Lin et~al.(2014)Lin, Maire, Belongie, Hays, Perona, Ramanan, Doll{\'a}r, and Zitnick]{lin2014microsoft}
T.-Y. Lin, M.~Maire, S.~Belongie, J.~Hays, P.~Perona, D.~Ramanan, P.~Doll{\'a}r, and C.~L. Zitnick.
\newblock Microsoft coco: Common objects in context.
\newblock In \emph{Computer Vision--ECCV 2014: 13th European Conference, Zurich, Switzerland, September 6-12, 2014, Proceedings, Part V 13}, pages 740--755. Springer, 2014.

\bibitem[Liu et~al.(2023)Liu, Emerson, and Collier]{liu2023visual}
F.~Liu, G.~Emerson, and N.~Collier.
\newblock Visual spatial reasoning.
\newblock \emph{Transactions of the Association for Computational Linguistics}, 11:\penalty0 635--651, 2023.

\bibitem[Luo et~al.(2023)Luo, Bao, Wu, He, and Li]{luo2023segclip}
H.~Luo, J.~Bao, Y.~Wu, X.~He, and T.~Li.
\newblock Segclip: Patch aggregation with learnable centers for open-vocabulary semantic segmentation.
\newblock In \emph{International Conference on Machine Learning}, pages 23033--23044. PMLR, 2023.

\bibitem[Ma et~al.(2023)Ma, Hong, Gul, Gandhi, Gao, and Krishna]{ma2023crepe}
Z.~Ma, J.~Hong, M.~O. Gul, M.~Gandhi, I.~Gao, and R.~Krishna.
\newblock Crepe: Can vision-language foundation models reason compositionally?
\newblock In \emph{Proceedings of the IEEE/CVF Conference on Computer Vision and Pattern Recognition}, pages 10910--10921, 2023.

\bibitem[Meng et~al.(2022)Meng, Bau, Andonian, and Belinkov]{p2}
K.~Meng, D.~Bau, A.~Andonian, and Y.~Belinkov.
\newblock Locating and editing factual associations in {GPT}.
\newblock In S.~Koyejo, S.~Mohamed, A.~Agarwal, D.~Belgrave, K.~Cho, and A.~Oh, editors, \emph{Advances in Neural Information Processing Systems 35: Annual Conference on Neural Information Processing Systems 2022, NeurIPS 2022, New Orleans, LA, USA, November 28 - December 9, 2022}, 2022.

\bibitem[Muennighoff et~al.(2023)Muennighoff, Tazi, Magne, and Reimers]{MTEB}
N.~Muennighoff, N.~Tazi, L.~Magne, and N.~Reimers.
\newblock {MTEB:} massive text embedding benchmark.
\newblock In A.~Vlachos and I.~Augenstein, editors, \emph{Proceedings of the 17th Conference of the European Chapter of the Association for Computational Linguistics, {EACL} 2023, Dubrovnik, Croatia, May 2-6, 2023}, pages 2006--2029. Association for Computational Linguistics, 2023.

\bibitem[Ni et~al.(2022)Ni, {\'{A}}brego, Constant, Ma, Hall, Cer, and Yang]{sent-t5}
J.~Ni, G.~H. {\'{A}}brego, N.~Constant, J.~Ma, K.~B. Hall, D.~Cer, and Y.~Yang.
\newblock Sentence-t5: Scalable sentence encoders from pre-trained text-to-text models.
\newblock In \emph{Findings of the Association for Computational Linguistics: {ACL} 2022}, pages 1864--1874. Association for Computational Linguistics, 2022.

\bibitem[Radford et~al.(2021)Radford, Kim, Hallacy, Ramesh, Goh, Agarwal, Sastry, Askell, Mishkin, Clark, et~al.]{radford2021learning}
A.~Radford, J.~W. Kim, C.~Hallacy, A.~Ramesh, G.~Goh, S.~Agarwal, G.~Sastry, A.~Askell, P.~Mishkin, J.~Clark, et~al.
\newblock Learning transferable visual models from natural language supervision.
\newblock In \emph{International conference on machine learning}, pages 8748--8763. PMLR, 2021.

\bibitem[Ramesh et~al.(2021)Ramesh, Pavlov, Goh, Gray, Voss, Radford, Chen, and Sutskever]{dalle}
A.~Ramesh, M.~Pavlov, G.~Goh, S.~Gray, C.~Voss, A.~Radford, M.~Chen, and I.~Sutskever.
\newblock Zero-shot text-to-image generation.
\newblock In \emph{International Conference on Machine Learning}, pages 8821--8831. PMLR, 2021.

\bibitem[Ramesh et~al.(2022)Ramesh, Dhariwal, Nichol, Chu, and Chen]{dalle-2}
A.~Ramesh, P.~Dhariwal, A.~Nichol, C.~Chu, and M.~Chen.
\newblock Hierarchical text-conditional image generation with clip latents.
\newblock \emph{arXiv preprint arXiv:2204.06125}, 2022.

\bibitem[Ray et~al.(2023)Ray, Radenovic, Dubey, Plummer, Krishna, and Saenko]{ray2023cola}
A.~Ray, F.~Radenovic, A.~Dubey, B.~A. Plummer, R.~Krishna, and K.~Saenko.
\newblock Cola: How to adapt vision-language models to compose objects localized with attributes?
\newblock \emph{arXiv preprint arXiv:2305.03689}, 2023.

\bibitem[Reimers and Gurevych(2019)]{sbert}
N.~Reimers and I.~Gurevych.
\newblock Sentence-bert: Sentence embeddings using siamese bert-networks.
\newblock In \emph{Proceedings of the 2019 Conference on Empirical Methods in Natural Language Processing}. Association for Computational Linguistics, 11 2019.

\bibitem[Reimers and Gurevych(2020)]{multi-lingual-clip}
N.~Reimers and I.~Gurevych.
\newblock Making monolingual sentence embeddings multilingual using knowledge distillation.
\newblock In B.~Webber, T.~Cohn, Y.~He, and Y.~Liu, editors, \emph{Proceedings of the 2020 Conference on Empirical Methods in Natural Language Processing, {EMNLP} 2020, Online, November 16-20, 2020}, pages 4512--4525. Association for Computational Linguistics, 2020.
\newblock \doi{10.18653/V1/2020.EMNLP-MAIN.365}.
\newblock URL \url{https://doi.org/10.18653/v1/2020.emnlp-main.365}.

\bibitem[Saharia et~al.(2022)Saharia, Chan, Saxena, Li, Whang, Denton, Ghasemipour, Gontijo~Lopes, Karagol~Ayan, Salimans, et~al.]{imagen}
C.~Saharia, W.~Chan, S.~Saxena, L.~Li, J.~Whang, E.~L. Denton, K.~Ghasemipour, R.~Gontijo~Lopes, B.~Karagol~Ayan, T.~Salimans, et~al.
\newblock Photorealistic text-to-image diffusion models with deep language understanding.
\newblock \emph{Advances in Neural Information Processing Systems}, 35:\penalty0 36479--36494, 2022.

\bibitem[Sahin et~al.(2024)Sahin, Li, Khan, Cremers, and Tresp]{sahin2024enhancing}
U.~Sahin, H.~Li, Q.~Khan, D.~Cremers, and V.~Tresp.
\newblock Enhancing multimodal compositional reasoning of visual language models with generative negative mining.
\newblock In \emph{Proceedings of the IEEE/CVF Winter Conference on Applications of Computer Vision}, pages 5563--5573, 2024.

\bibitem[Schiappa et~al.(2023)Schiappa, Cogswell, Divakaran, and Rawat]{schiappa2023probing}
M.~C. Schiappa, M.~Cogswell, A.~Divakaran, and Y.~S. Rawat.
\newblock Probing conceptual understanding of large visual-language models.
\newblock \emph{arXiv preprint arXiv:2304.03659}, 2023.

\bibitem[Schuhmann et~al.(2022)Schuhmann, Beaumont, Vencu, Gordon, Wightman, Cherti, Coombes, Katta, Mullis, Wortsman, et~al.]{schuhmann2022laion}
C.~Schuhmann, R.~Beaumont, R.~Vencu, C.~Gordon, R.~Wightman, M.~Cherti, T.~Coombes, A.~Katta, C.~Mullis, M.~Wortsman, et~al.
\newblock Laion-5b: An open large-scale dataset for training next generation image-text models.
\newblock \emph{Advances in Neural Information Processing Systems}, 35:\penalty0 25278--25294, 2022.

\bibitem[Singh et~al.(2022)Singh, Hu, Goswami, Couairon, Galuba, Rohrbach, and Kiela]{singh2022flava}
A.~Singh, R.~Hu, V.~Goswami, G.~Couairon, W.~Galuba, M.~Rohrbach, and D.~Kiela.
\newblock Flava: A foundational language and vision alignment model.
\newblock In \emph{Proceedings of the IEEE/CVF Conference on Computer Vision and Pattern Recognition}, pages 15638--15650, 2022.

\bibitem[Singh et~al.(2023)Singh, Zhang, Wang, Wang, Xiong, Du, and Chen]{SinghZWWXDC23}
H.~Singh, P.~Zhang, Q.~Wang, M.~Wang, W.~Xiong, J.~Du, and Y.~Chen.
\newblock Coarse-to-fine contrastive learning in image-text-graph space for improved vision-language compositionality.
\newblock In \emph{Proceedings of the 2023 Conference on Empirical Methods in Natural Language Processing, {EMNLP} 2023}, pages 869--893. Association for Computational Linguistics, 2023.

\bibitem[Su et~al.(2023)Su, Shi, Kasai, Wang, Hu, Ostendorf, Yih, Smith, Zettlemoyer, and Yu]{INSTRUCTOR}
H.~Su, W.~Shi, J.~Kasai, Y.~Wang, Y.~Hu, M.~Ostendorf, W.~Yih, N.~A. Smith, L.~Zettlemoyer, and T.~Yu.
\newblock One embedder, any task: Instruction-finetuned text embeddings.
\newblock In \emph{Findings of the Association for Computational Linguistics: {ACL} 2023}, pages 1102--1121. Association for Computational Linguistics, 2023.

\bibitem[Taladngoen and Esteban(2022)]{taladngoen2022assumptions}
U.~Taladngoen and R.~H. Esteban.
\newblock Assumptions on plausible lexical distractors in the redesigned toeic question-response listening test.
\newblock \emph{LEARN Journal: Language Education and Acquisition Research Network}, 15\penalty0 (2):\penalty0 802--829, 2022.

\bibitem[Thrush et~al.(2022)Thrush, Jiang, Bartolo, Singh, Williams, Kiela, and Ross]{thrush2022winoground}
T.~Thrush, R.~Jiang, M.~Bartolo, A.~Singh, A.~Williams, D.~Kiela, and C.~Ross.
\newblock Winoground: Probing vision and language models for visio-linguistic compositionality.
\newblock In \emph{Proceedings of the IEEE/CVF Conference on Computer Vision and Pattern Recognition}, pages 5238--5248, 2022.

\bibitem[Touvron et~al.(2023)Touvron, Lavril, Izacard, Martinet, Lachaux, Lacroix, Rozi{\`e}re, Goyal, Hambro, Azhar, et~al.]{touvron2023llama}
H.~Touvron, T.~Lavril, G.~Izacard, X.~Martinet, M.-A. Lachaux, T.~Lacroix, B.~Rozi{\`e}re, N.~Goyal, E.~Hambro, F.~Azhar, et~al.
\newblock Llama: Open and efficient foundation language models.
\newblock \emph{arXiv preprint arXiv:2302.13971}, 2023.

\bibitem[Wang et~al.(2019)Wang, Singh, Michael, Hill, Levy, and Bowman]{WangSMHLB19}
A.~Wang, A.~Singh, J.~Michael, F.~Hill, O.~Levy, and S.~R. Bowman.
\newblock {GLUE:} {A} multi-task benchmark and analysis platform for natural language understanding.
\newblock In \emph{7th International Conference on Learning Representations, {ICLR} 2019}. OpenReview.net, 2019.

\bibitem[Wang et~al.(2023{\natexlab{a}})Wang, Ding, Rao, Liu, Shen, and Ding]{wang2023can}
F.~Wang, L.~Ding, J.~Rao, Y.~Liu, L.~Shen, and C.~Ding.
\newblock Can linguistic knowledge improve multimodal alignment in vision-language pretraining?
\newblock \emph{arXiv preprint arXiv:2308.12898}, 2023{\natexlab{a}}.

\bibitem[Wang et~al.(2022)Wang, Yang, Huang, Jiao, Yang, Jiang, Majumder, and Wei]{wang2022text}
L.~Wang, N.~Yang, X.~Huang, B.~Jiao, L.~Yang, D.~Jiang, R.~Majumder, and F.~Wei.
\newblock Text embeddings by weakly-supervised contrastive pre-training.
\newblock \emph{arXiv preprint arXiv:2212.03533}, 2022.

\bibitem[Wang et~al.(2023{\natexlab{b}})Wang, Yang, Huang, Yang, Majumder, and Wei]{wang2023improving}
L.~Wang, N.~Yang, X.~Huang, L.~Yang, R.~Majumder, and F.~Wei.
\newblock Improving text embeddings with large language models.
\newblock \emph{arXiv preprint arXiv:2401.00368}, 2023{\natexlab{b}}.

\bibitem[Wang et~al.(2020)Wang, Wei, Dong, Bao, Yang, and Zhou]{wang2020minilm}
W.~Wang, F.~Wei, L.~Dong, H.~Bao, N.~Yang, and M.~Zhou.
\newblock Minilm: Deep self-attention distillation for task-agnostic compression of pre-trained transformers, 2020.

\bibitem[Xiao et~al.(2023)Xiao, Liu, Zhang, and Muennighoff]{bge_embedding}
S.~Xiao, Z.~Liu, P.~Zhang, and N.~Muennighoff.
\newblock C-pack: Packaged resources to advance general chinese embedding, 2023.

\bibitem[Xu et~al.(2023)Xu, Shao, Zhang, Gao, Liu, Lei, Meng, Huang, Qiao, and Luo]{xu2023lvlm}
P.~Xu, W.~Shao, K.~Zhang, P.~Gao, S.~Liu, M.~Lei, F.~Meng, S.~Huang, Y.~Qiao, and P.~Luo.
\newblock Lvlm-ehub: A comprehensive evaluation benchmark for large vision-language models.
\newblock \emph{arXiv preprint arXiv:2306.09265}, 2023.

\bibitem[Yang et~al.(2023)Yang, Deng, An, Li, Feng, Guo, Yang, and Liu]{yang2023alip}
K.~Yang, J.~Deng, X.~An, J.~Li, Z.~Feng, J.~Guo, J.~Yang, and T.~Liu.
\newblock Alip: Adaptive language-image pre-training with synthetic caption.
\newblock In \emph{Proceedings of the IEEE/CVF International Conference on Computer Vision}, pages 2922--2931, 2023.

\bibitem[Yuksekgonul et~al.(2023)Yuksekgonul, Bianchi, Kalluri, Jurafsky, and Zou]{yuksekgonul2023and}
M.~Yuksekgonul, F.~Bianchi, P.~Kalluri, D.~Jurafsky, and J.~Zou.
\newblock When and why vision-language models behave like bags-of-words, and what to do about it?
\newblock In \emph{The Eleventh International Conference on Learning Representations}, 2023.

\bibitem[Zeng et~al.(2022)Zeng, Zhang, and Li]{zeng2022multi}
Y.~Zeng, X.~Zhang, and H.~Li.
\newblock Multi-grained vision language pre-training: Aligning texts with visual concepts.
\newblock In \emph{International Conference on Machine Learning}, pages 25994--26009. PMLR, 2022.

\bibitem[Zeng et~al.(2023)Zeng, Yu, Gao, Meng, Goyal, and Chen]{eval-ins-following}
Z.~Zeng, J.~Yu, T.~Gao, Y.~Meng, T.~Goyal, and D.~Chen.
\newblock Evaluating large language models at evaluating instruction following.
\newblock \emph{CoRR}, abs/2310.07641, 2023.

\bibitem[Zhao et~al.(2022)Zhao, Zhang, Zhu, Shen, Lee, Lu, and Yin]{zhao2022vl}
T.~Zhao, T.~Zhang, M.~Zhu, H.~Shen, K.~Lee, X.~Lu, and J.~Yin.
\newblock Vl-checklist: Evaluating pre-trained vision-language models with objects, attributes and relations.
\newblock \emph{arXiv preprint arXiv:2207.00221}, 2022.

\bibitem[Zheng et~al.(2023)Zheng, Chiang, Sheng, Zhuang, Wu, Zhuang, Lin, Li, Li, Xing, Zhang, Gonzalez, and Stoica]{llm-judge}
L.~Zheng, W.~Chiang, Y.~Sheng, S.~Zhuang, Z.~Wu, Y.~Zhuang, Z.~Lin, Z.~Li, D.~Li, E.~P. Xing, H.~Zhang, J.~E. Gonzalez, and I.~Stoica.
\newblock Judging llm-as-a-judge with mt-bench and chatbot arena.
\newblock \emph{CoRR}, abs/2306.05685, 2023.

\bibitem[Zhou et~al.(2023{\natexlab{a}})Zhou, Lai, Yeong, Mouratidis, and Jiang]{rome_dataset}
K.~Zhou, E.~Lai, W.~B.~A. Yeong, K.~Mouratidis, and J.~Jiang.
\newblock {ROME:} evaluating pre-trained vision-language models on reasoning beyond visual common sense.
\newblock In \emph{Findings of the Association for Computational Linguistics: {EMNLP} 2023}, pages 10185--10197. Association for Computational Linguistics, 2023{\natexlab{a}}.

\bibitem[Zhou et~al.(2023{\natexlab{b}})Zhou, Muresanu, Han, Paster, Pitis, Chan, and Ba]{zhou2023large}
Y.~Zhou, A.~I. Muresanu, Z.~Han, K.~Paster, S.~Pitis, H.~Chan, and J.~Ba.
\newblock Large language models are human-level prompt engineers.
\newblock In \emph{The Eleventh International Conference on Learning Representations}, 2023{\natexlab{b}}.
\newblock URL \url{https://openreview.net/forum?id=92gvk82DE-}.

\bibitem[Zou et~al.(2023)Zou, Wang, Kolter, and Fredrikson]{p1}
A.~Zou, Z.~Wang, J.~Z. Kolter, and M.~Fredrikson.
\newblock Universal and transferable adversarial attacks on aligned language models.
\newblock \emph{CoRR}, abs/2307.15043, 2023.
\newblock \doi{10.48550/ARXIV.2307.15043}.
\newblock URL \url{https://doi.org/10.48550/arXiv.2307.15043}.

\end{thebibliography}

\newpage
\appendix
\onecolumn



\section{VISLA Benchmark Generation}
\subsection{Dataset Guidelines}
\label{guidelines}
The main guidelines followed to create the benchmarks are:
\begin{itemize}
\item The lexical changes allowed to creation the three captions, include replacing words with synonyms and antonyms, reordering the words, etc. These lexical changes do not include adding more details about the image in the caption;
\item Due to the lexical alterations, the three captions should not consist of any nonsensical and non-fluent errors;
\item The three captions should be generated such that they do not need any visual, logical or commonsense reasoning to distinguish the semantically similar captions ($C_{P1}$ and $C_{P2}$) from the semantically different caption ($C_N$) i.e., given only three captions without image, one should be able to distinguish $C_{P1}$, $C_{P2}$ from $C_{N}$. 

\item To ensure fairness and avoid bias in dataset, we used gender neutral words such as 'person', 'individual', etc instead of using gender specific pronouns such as he, she, him , her etc.
\end{itemize}


\subsection{Prompt for generic \task dataset}
\label{appendix:generic-data-prompt}
\begin{figure*}[htb!]
    \centering
    
    \begin{tcolorbox}[
        colback=gray!10, 
        colframe=gray,
        arc=5mm,
        boxrule=1pt,
    ]
    \textbf{Rules Instruction:} Given an input sentence describing an image caption, follow these steps:
    \begin{enumerate}
        \item Rephrase each provided sentence, focusing on preserving the original spatial relationship.
        \item Pay careful attention to the positioning of objects or entities in relation to one another.
        \item Ensure that the meaning remains consistent and that both the original and paraphrased sentences maintain logical coherence and grammatical correctness.
    \end{enumerate}
    \textbf{Demonstration:} For example,
    
    \textbf{\graybox{Input:}} Cat is under the table. \\
    \textbf{\graybox{Paraphrase Idea:}} Rephrase the sentence to convey that the table is positioned above the cat.\\
     \textbf{\graybox{Paraphrased:}} The table is above the cat.
    
    Another example,
    
    \textbf{\graybox{Input:}}The plane flies below the bright white clouds.\\
    \textbf{\graybox{Paraphrase Idea:}} Ensure the spatial context is maintained by stating that the bright white clouds are situated above the flying plane.\\
     \textbf{\graybox{Paraphrased:}} The plane flies below the bright white clouds.
    
    Similarly,
    
    \textbf{\graybox{Input:}} The third balcony is below the fourth balcony from the bottom.\\
    \textbf{\graybox{Paraphrase Idea:}} Emphasize the consistent spatial arrangement while indicating that the fourth balcony is positioned above the third balcony from the bottom.\\
     \textbf{\graybox{Paraphrased:}} The fourth balcony is above the third balcony from the bottom.\\

    \textit{Remember to keep the meaning intact, and both the original and paraphrased sentences should be logically coherent and grammatically correct.} \\

    Lastly, for the final example:\\
    \textbf{\graybox{Input:}} \textit{[Original caption goes here]} \\
    \textbf{\graybox{Paraphrase Idea:}} Focus on replicating the spatial arrangement while maintaining the original meaning of the sentence, correct grammar, same meaning. \\
    \textbf{\graybox{Paraphrased:}} \textit{[Your paraphrased sentence goes here]}
    \end{tcolorbox}
    \vspace{-8pt}
    \caption{Rules Prompt used for priming LLM after role-playing instructions.}
    \label{fig:prompting}
\end{figure*}

\newpage
\subsection{Validation prompt for Generic \task dataset}
\label{app:val_prompt}
Figure \ref{fig:LLM-evaluation} shows the comprehensive prompt used to validate the samples generated by priming the LLM. The outputs obtained from this prompt are further validated by a human expert. This reduces the manual effort required to create the \task benchmark.
\begin{figure*}[ht]
    \centering
    \begin{tcolorbox}[
        colback=gray!10, 
        colframe=gray,
        arc=5mm,
        boxrule=1pt,
    ]
    \textbf{Instruction:}Given a pair of captions you job is to check if the second caption is consistent with the first caption.
    If it is consistent output the second caption as is, Otherwise rephrase the second caption to be consistent with the first sentence.
    We are especially interested in spatial consistency and spatial relationship of the objects with each other.\\
    \textbf{Demonstrations:} examples, \\
    \textbf{\graybox{Caption 1:}}  A guy holding a skateboard is speaking into a microphone.\\
    \textbf{\graybox{Caption 2:}}  The guy holding the microphone is speaking into the skateboard.\\
    \textbf{\graybox{isConsistent}}: No, you cannot speak into a skateboard.\\
    \textbf{\graybox{newCaption}}: The guy is speaking into the microphone while holding a skateboard. \\
    
    \textbf{\graybox{Caption 1:}} A family are playing frisbee on the beach.\\
    \textbf{\graybox{Caption 2:}} The frisbee is being played on the beach by a family.\\
    \textbf{\graybox{isConsistent}}: Yes, caption 2 is consistent as it is the same caption written in passive voice. new caption is the same as caption 2.\\
    \textbf{\graybox{newCaption}}: A family are playing frisbee on the beach.\\
    
     \textbf{\graybox{caption 1:}}A stop sign vandalized with an "eating animals" sticker below the word "stop."\\
    \textbf{\graybox{caption 2:}} The stop sign is below an "eating animals" sticker.\\
    \textbf{\graybox{isConsistent}}: The stop cannot be below and above the sticker at the same time.\\
   \textbf{\graybox{newCaption}}: The word "stop" sign is above an "eating animals" sticker.\\
    
     \textbf{\graybox{caption 1:}}There is a phone on top of a calculator.\\
    \textbf{\graybox{caption 2:}} A calculator lies beneath the phone.\\
     \textbf{\graybox{isConsistent}}:Yes, the sentences are semantically equivalent. new caption is same as caption 2.\\
    \textbf{\graybox{newCaption}}: A calculator lies beneath the phone.\\
    
    Now the same for the below caption only.\\
    \textbf{\graybox{caption 1:}} \textit{[Original caption goes here]} \\
   \textbf{\graybox{caption 2:}} \textit{[Generated caption goes here]} \\
    \textbf{\graybox{isConsistent}}: \textit{[Output Here]}
    
    \end{tcolorbox}
    \caption{LLM Validation prompt to evaluate the generated caption.}
    \label{fig:LLM-evaluation}
\end{figure*}
\newpage

\subsection{Samples from generic dataset}
\label{appendix:generic_examples}
\begin{figure}[htb]
    \centering
    \vspace{0.5cm}
    \includegraphics[width=1.0\linewidth]{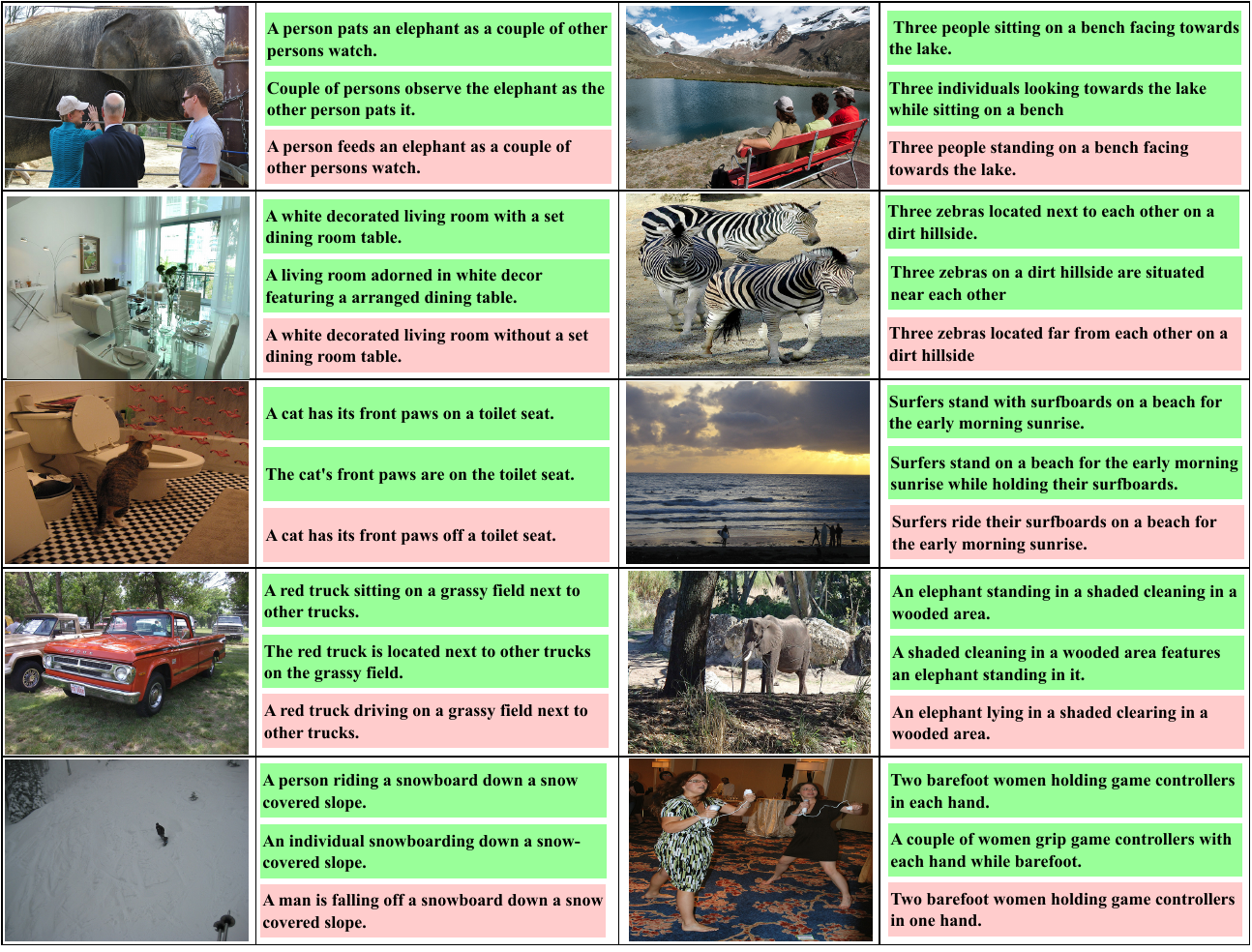}
    \caption{\small Some samples from the \task generic evaluation dataset.}
    \label{fig:generic_ex}
\end{figure}

Figure \ref{fig:generic_ex} shows examples from the generic \task dataset. Lexical alterations such as synonyms/antonyms of words, negations, re-ordering the words and adding non-content words were used to generate the semantically equivalent pair, and the semantically different negative caption. Most of the sentences in the dataset are generated by using combination of multiple lexical alterations.

\newpage
\subsection{Samples from spatial dataset}
\label{spatial_ex}
\begin{figure}[htb]
    \centering
    \vspace{0.5cm}
    \includegraphics[width=1.0\linewidth]{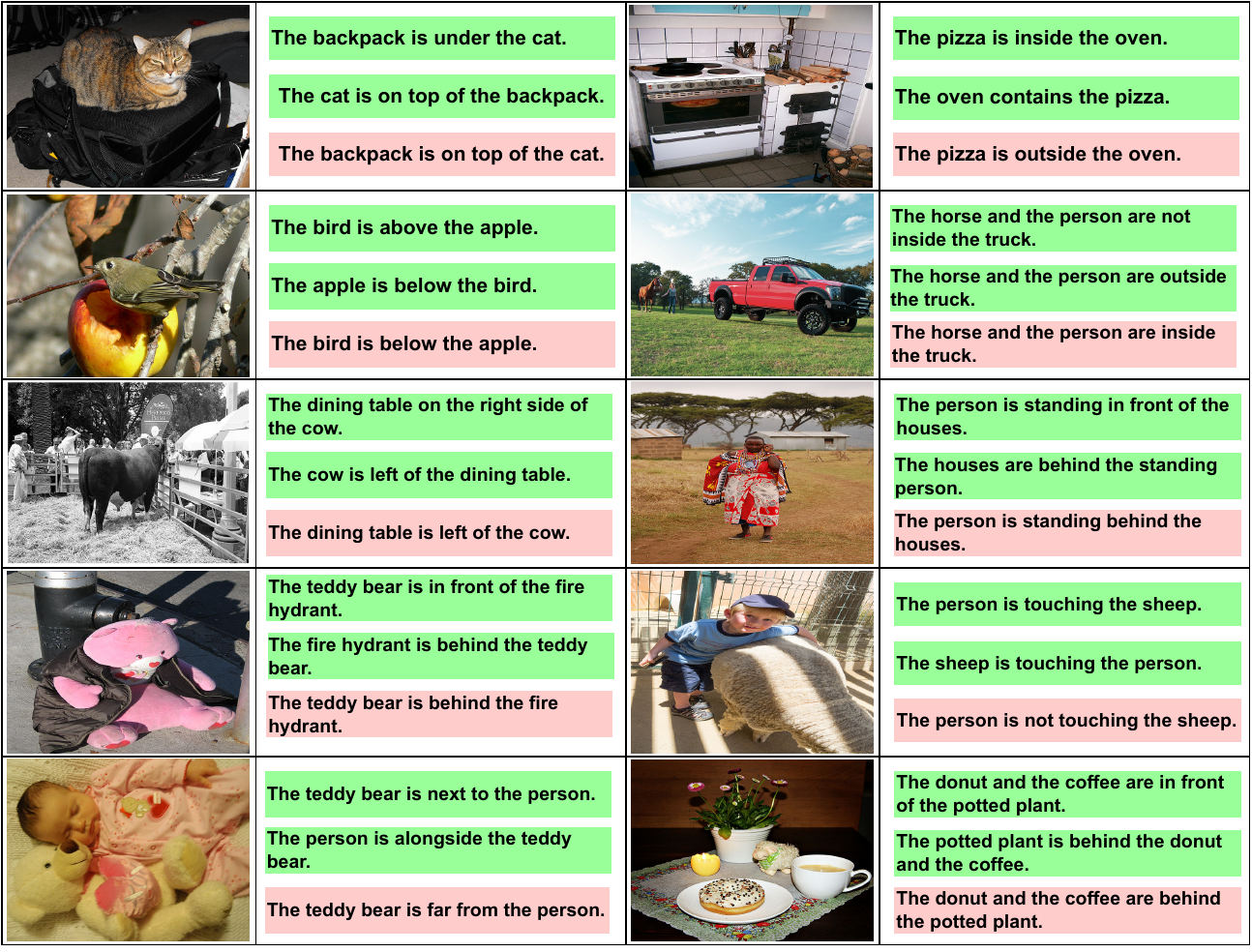}
    \caption{\small Some samples from the \task spatial evaluation dataset.}
    \label{fig:spatial_ex}
\end{figure}

Figure \ref{fig:spatial_ex} shows examples from the spatial \task dataset. This dataset mainly focus on the spatial arrangement of objects in the images. The semantically equivalent pair and the semantically different negative caption are generated by using different types of lexical alterations such as synonyms and antonyms of words, negations, re-ordering the words and swapping the subject and the object. Most of the sentences in the dataset are generated by using multiple lexical alterations.

\newpage
\section{Retrieval settings}
\label{appendix:retrival-settings}

We use two retrieval settings for evaluation in the \task task as described below,
\paragraph{Text-to-Text Retrieval (T2T):} We assess the unimodal and multimodal text encoder by providing the triplet caption set as input. For the T2T task, we extract text embeddings for the positive Caption 1 ($E_{P1}$), positive Caption 2 ($E_{P2}$), and negative Caption ($E_{N}$). We compute pairwise cosine similarity scores between Caption 1, Caption 2 and negative Caption
($S(E_{P1}, E_{P2})$, $S(E_{P1},E_{N})$, and $S(E_{P2},E_{N})$). 
The values of these cosine similarities determine the rank of the positive and the negative captions. We report the accuracy of the model, assigning the last rank to the negative captions.
\vspace{-5pt}
\paragraph{Image-to-Text Retrieval (I2T):} In this setting, we provide the image and the corresponding caption triplets as input to the Vision Language Models (VLMs). For the I2T task, we extract the image embedding ($E_I$), text embeddings for each positive Caption 1 ($E_{P1}$), positive Caption 2 ($E_{P2}$), and negative Caption ($E_{N}$). We compute cosine similarity scores between the image embedding and each text embedding ($S(E_I, E_{P1})$, $S(E_I, E_{P2})$, and $S(E_I,E_N)$). Similar to the T2T task, We report the accuracy of the model, assigning the last rank to the negative captions.

\section{Additional Analysis}
\subsection{Lexical overlap of captions}
\label{appendix:lexical_overlap}
Figure 8, demonstrates the intentional lexical overlap of semantically close caption pairs in \task. We employs edit distance as a measure for lexical overlap, showcasing the controlled setting used to investigate the sensitivity of embeddings to lexical versus semantic changes in language models.
\begin{figure*}[!htbp]
     \centering
     \begin{subfigure}[b]{0.48\linewidth}
         \centering
         \includegraphics[width=1\textwidth]{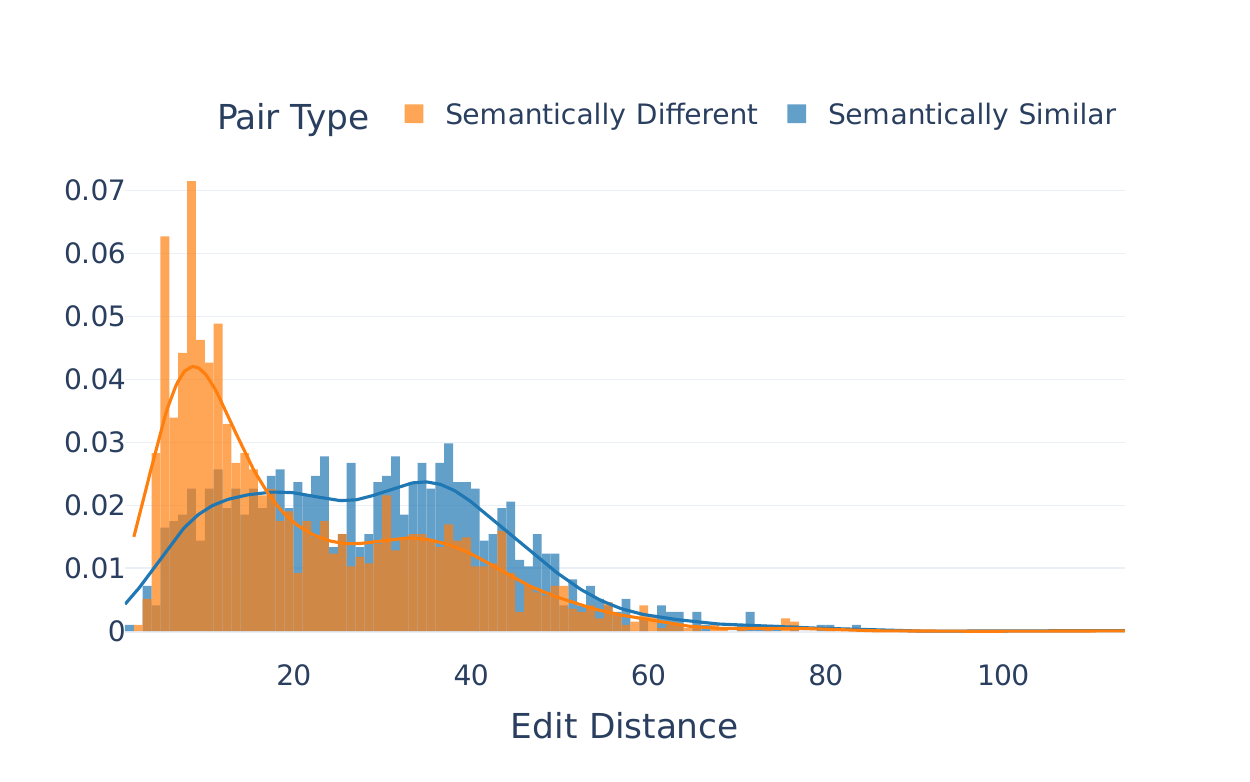}
         \caption{Generic \task dataset}
         \label{fig:dist_comb_generic}
     \end{subfigure}
     \hfill
     \begin{subfigure}[b]{0.48\linewidth}
         \centering
         \includegraphics[width=1\textwidth]{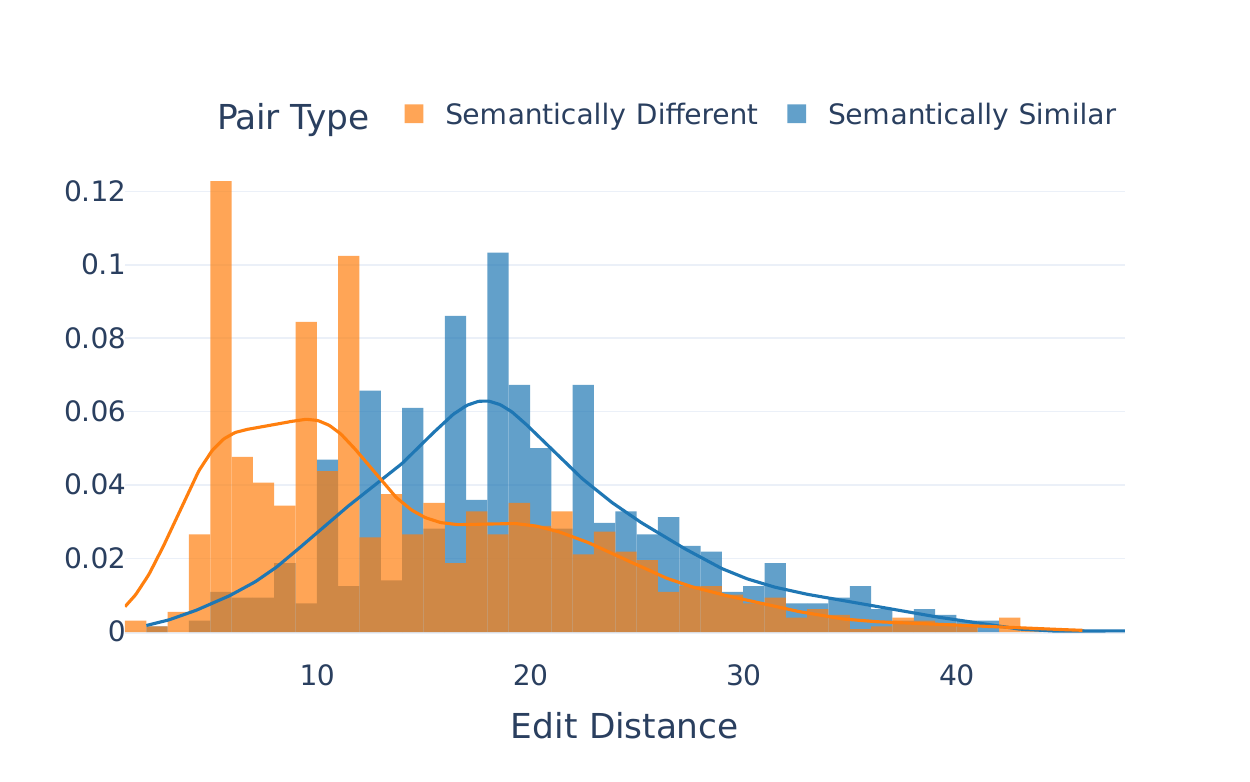}
         \caption{Spatial \task dataset}
         \label{fig:dist_comb_spatial}
     \end{subfigure}
        \caption{Distribution of edit distances between semantically similar pairs (i.e., among positive captions) and different pairs (i.e., between positive and negative captions) in the spatial \task dataset. Semantically similar sentences exhibit higher edit distances, indicating lexical differences. In contrast, semantically different sentences have lower edit distances, suggesting lexical similarities.}
        \label{fig:dist_comb_labs}
\end{figure*}

\newpage
\subsection{Detailed Results of ULMs on \task}
\label{appendix:det_ulm}
\begin{table}[!htbp]
    \centering
    \caption{\small Comparison of ULMs on the Generic \task and the Spatial \task dataset. P1-N and P2-N refer to the accuracy (\%) of ranking positive caption 1 and positive caption 2 above the Negative caption, respectively. P1- captions have more lexical overlap with the negative caption.  Best performance within the same scale category is underlined, and across all models is bold-faced.We include the number of parameters in text encoders relative to BERT-base, i.e., 109.5
million parameters.}
        \label{tab:unimodal-performance-finegrained_full}
        \vspace{0.5cm}
    \resizebox{\linewidth}{!}{
        \begin{tabular}{lllllll}
            \toprule
            \textbf{Dataset} &  \textbf{Dim}&\textbf{\# Params}&\multicolumn{2}{c}{\textbf{Generic}}& \multicolumn{2}{c}{\textbf{Spatial}}\\ 
            \cmidrule(lr){4-5} \cmidrule(lr){6-7}
            \textbf{Model} &    &(BERT Scale) &\textbf{P1-N Acc(\%)}& \textbf{P2-N Acc(\%)}& \textbf{ P1-N Acc(\%)}&\textbf{P2-N Acc(\%)}\\
            \toprule
            All-MiniLM-L6-v2 \cite{wang2020minilm} &   384&0.21 &93.22&60.12&\underline{54.38}&46.56\\ 
            BGE-small-en-v1.5 \cite{bge_embedding}&   384&0.3 &93.73&\underline{73.48}&53.28&\underline{52.81}\\ 
            All-MiniLM-L12-v2 \cite{wang2020minilm} &   384&0.3 &\underline{94.35}&66.91&53.75&50.31\\
            GTE-small \cite{li2023general} &   384&0.3 &\underline{94.66}&64.85&\underline{54.69}&\underline{52.19}\\
            \midrule
            Angle-BERT-base-uncased-nli-en-v1 \cite{li2023angle}  &   768&1  &\underline{94.66}&\underline{74.31}&\underline{55.94}&\underline{58.28}\\ 
            BGE-base-en-v1.5  \cite{bge_embedding} &   768&1 &\underline{94.14}&73.38&54.38&55.78\\ 
            Sentence-T5-base \cite{sent-t5} &   768&1.01  &\underline{94.24}&72.05&\underline{55}&\underline{58.59}\\ 
            GTE-base \cite{li2023general} &   768&1 &\underline{94.55}&67.73&\underline{55.63}&54.22\\ 
            Clip-ViT-B-32-multilingual-v1 \cite{multi-lingual-clip} &   512&1.23 &78.83&40.90&52.97&45.47\\ 
            Clip-ViT-B-32 \cite{radford2021learning} &   512&1.38 &81.71&35.46&52.66&35.78\\
            \midrule
            Instructor-large \cite{INSTRUCTOR} &   768&3.07 &93.63&73.48&\textbf{56.72}&57.66\\ 
            Instructor-large(custom-ins)\cite{INSTRUCTOR} &   768&3.07 &\underline{94.86}&\underline{75.64}&\textbf{56.09}&61.25\\ 
            UAE-Large-V1 \cite{li2023angle} &   1024&3.06 &\underline{94.14}&74.20&54.69&58.59\\  
            GTE-large \cite{li2023general} &   1024&3.06 &\underline{94.76}&68.65&55.16&57.66\\ 
            All-RoBERTa-large-v1 \cite{sbert}  &   1024&3.25 &93.94&72.97&55&52.5\\ 
            Stsb-RoBERTa-large \cite{sbert} &   1024&3.25 &93.01&71.63&54.22&\textbf{66.88}\\ 
            LaBSE \cite{labse} &   768&4.31 &84.89&35.05&53.28&45.94\\ 
            Sentence-T5-xl \cite{sent-t5} &   768&11.34 &\underline{94.55}&72.66&55.31&63.75\\
            \midrule
            Angle-Llama-7b-nli-v2 \cite{li2023angle}&   4096&62.28 &\textbf{95.89}&\textbf{79.34}&\textbf{56.41}&\underline{61.41}\\
            E5-Mistral-7b-instruct \cite{wang2023improving, wang2022text} &   4096&64.95 &\textbf{95.58}&78.93&55.31&60.16\\
            \bottomrule
        \end{tabular}}
    \end{table}

\subsection{Specification of VLMs evaluated on \task}
\label{appendix:vlm_eval}
We comprehensively evaluate a wide array of VLMs, which include:

1) Models trained with a contrastive learning objective such as CLIP-ViT-B/32~\cite{radford2021learning}, RoBERTa-ViT-B/32~\cite{schuhmann2022laion}, ALIGN~\cite{jia2021scaling} and ALIP~\cite{yang2023alip}. ALIGN and ALIP utilize noisy and synthetic captions, respectively. 

2) Models trained by combining multiple objective functions, such as FLAVA~\cite{singh2022flava}: pretrained by combining contrastive, Image-text matching (ITM), masked image modeling (MIM) and masked language modeling (MLM) objectives;  ALBEF~\cite{li2021albef}: which combines ITM and MLM; BLIP~\cite{li2022blip} and BLIP-2~\cite{li2023blip2}: which combine contrastive, ITM and image captioning objectives.

3) Models with a unified encoder for text and images, such as ViLT~\cite{kim2021vilt}, and multi-lingual distilled models like AltCLIP~\cite{ChenLZYW23} 

4) Models that align text with corresponding visual concepts in the image, such as SegCLIP~\cite{luo2023segclip}, and XVLM~\cite{zeng2022multi} - with two variants, XVLM-4M and XVLM-16M.

We also investigate several models that have been finetuned on downstream tasks of image-text retrieval, such as BLIP-ITM-COCO~\cite{li2022blip}, ViLT-ITR-COCO \cite{kim2021vilt} and XVLM-16M-ITR-COCO \cite{zeng2022multi}. Specifically, BLIP, ViLT, and XVLM-16M models were trained for the ITM task using the COCO dataset. Additionally, XVLM-16M-ITR-Flickr \cite{zeng2022multi} denotes XVLM-16M models trained for the ITM task using the Flickr dataset. 

Moreover, we evaluate recent methods proposed to improve the compositionality of VLMs, including NegCLIP \cite{yuksekgonul2023and}, SVLC \cite{doveh2023teaching}, CyCLIP \cite{goel2022cyclip}, and BLIP-SGVL \cite{HerzigMKAFDG23}. 

Table \ref{tab:vlm_details} provide further details about different VLMs.


\begin{table}[htb!]
\vspace{-5cm}
\caption{Details of the VLMs evaluated using the VISLA benchmarks. Pretraining Data type: R, N and S refer to Real, Noisy and Synthetic data types, respectively. Pretraining Objectives -- ITC: image-text contrastive; ITM: image-text matching; MLM: masked language modeling; MMM: masked multimodal modeling; MIM: masked image modeling; IC: image captioning; IS: image segmentation using KL divergence; ITA: image-text alignment; CCL: Cycle-consistency loss; finetuning objectives -- ITR: image-text retrieval; NL: Negative loss for text; SG: scene graph loss; PT, FT refer to pretraining and finetuning, respectively}
\label{tab:vlm_details}
\vspace{0.5cm}
\begin{tabular}{l|lllllll}
\toprule
\multirow{2}{*}{Model} & \#Total & Embedding & Pretraining & Pretraining & Pretraining & \multirow{2}{*}{Finetuned} \\
& Parameters & Dimension & Data size & Data Type & Objectives &  \\
\midrule 
CLIP-ViT-B-32  \citeyear{radford2021learning} & 151M & 512 & 400M & R & ITC & \xmark \\
RoBERTa-ViT-B-32 \citeyear{schuhmann2022laion} & 212M & 512 & 2B & R & ITC &  \xmark \\
ALIGN \citeyear{jia2021scaling} & 490M & 640 & 1.8B & R+N & ITC & \xmark \\
ALIP \citeyear{yang2023alip}  & 151M & 512 & 15M & R+S & ITC & \xmark \\
\midrule
\multirow{2}{*}{FLAVA \citeyear{singh2022flava}}  & \multirow{2}{*}{358M} & \multirow{2}{*}{768} & \multirow{2}{*}{70M} & \multirow{2}{*}{R} & ITC, ITM, MLM & \multirow{2}{*}{\xmark} \\
 & & & & & MMM, MIM & \\
 \hdashline
ALBEF \citeyear{li2021albef} & 210M & 256 & 14M & R+N & ITC, ITM, MLM & \xmark  \\
BLIP \citeyear{li2022blip}  & 225M & 512 & 129M & R+S & ITC, ITM, IC & \xmark \\
BLIP2 \citeyear{li2023blip2}  & 1173M & 256 & 129M & R+S & ITC, ITM, IC & \xmark  \\
\midrule
ViLT \citeyear{kim2021vilt}  & 111M & 768 & 10M & R & ITM, MLM & \xmark  \\
AltCLIP \citeyear{ChenLZYW23}  & 864M & 768 & 42M & R & ITC & \xmark  \\
\midrule
SegCLIP \citeyear{luo2023segclip} & 151M & 512 & 400M+4M & R & ITC, MIM, IS & \xmark \\
XVLM-4M \citeyear{zeng2022multi} & 216M & 256 & 4M & R & ITC, ITM, MLM, ITA & \xmark \\
XVLM-16M \citeyear{zeng2022multi} & 216M & 256 & 16M & R & ITC, ITM, MLM, ITA & \xmark \\
\midrule
\multirow{2}{*}{BLIP-ITM-COCO \citeyear{li2022blip}}  & \multirow{2}{*}{223M} & \multirow{2}{*}{512} & PT: 129M & R+S & ITC, ITM, IC & \multirow{2}{*}{\cmark} \\
&&& FT: 110K & R & FT: ITM & \\
\hdashline
\multirow{2}{*}{ViLT-ITR-COCO \citeyear{kim2021vilt}}  & \multirow{2}{*}{111M} & \multirow{2}{*}{768} & PT: 10M & \multirow{2}{*}{R} & ITM, MLM & \multirow{2}{*}{\cmark} \\
&&& FT: 110K &  & FT: ITR & \\
\hdashline
\multirow{2}{*}{XVLM-16M-COCO \citeyear{zeng2022multi}} & \multirow{2}{*}{216M} & \multirow{2}{*}{256} & PT:16M & \multirow{2}{*}{R} & ITC, ITM, MLM, ITA & \multirow{2}{*}{\cmark}  \\
&&& FT: 110K &  & FT: ITR & \\
\hdashline
\multirow{2}{*}{XVLM-16M-Flickr \citeyear{zeng2022multi}} & \multirow{2}{*}{216M} & \multirow{2}{*}{256} & PT: 16M & \multirow{2}{*}{R} & TC, ITM, MLM, ITA & \multirow{2}{*}{\cmark}  \\
&&& FT: 30K &  & FT: ITR & \\
\midrule
\multirow{2}{*}{NegCLIP \citeyear{yuksekgonul2023and}}  & \multirow{2}{*}{151M} & 512 & PT: 400M & \multirow{2}{*}{R} & ITC & \multirow{2}{*}{\cmark} \\
& & & FT:110K &  & FT: ITM & \\
\hdashline
\multirow{2}{*}{CLIP-SVLC \citeyear{doveh2023teaching}}  & \multirow{2}{*}{151M} & \multirow{2}{*}{512} & PT:400M & \multirow{2}{*}{R} & ITC & \multirow{2}{*}{\cmark}  \\
& & & FT:400M &  & FT: ITC, NL & \\
\hdashline
\multirow{2}{*}{BLIP-SGVL \citeyear{HerzigMKAFDG23}}  & \multirow{2}{*}{696M} & \multirow{2}{*}{768} & PT: 129M & \multirow{2}{*}{R} & ITC, ITM, IC & \multirow{2}{*}{\cmark} \\
& & & FT:4M &  & FT: ITC, SG & \\
\hdashline
CyCLIP \citeyear{goel2022cyclip}  & 102M & 1024 & PT: 102M & R & ITC, CCL &  \xmark \\
\bottomrule
\end{tabular}
\end{table}

\clearpage
\subsection{CLIP variants evaluation on \task}
\label{app:clip_var}
\begin{table*}[htb!]
\caption{\small Comparison between the performance of different variants of CLIP when tested on the generic and spatial \task benchmarks. Data, Model and Emb. refer to the pre-training dataset size and total number of parameters in the model (in Millions) and embedding dimension, respectively. Performance reported in terms of Accuracy (\%)}
\label{tab:clip_variants}
\vspace{0.5cm}
\centering
\begin{tabular}{l|lrrr|cccc}
\toprule
& Pre-training& Pre-training & \# Params & Embed. & \multicolumn{2}{c}{Generic} & \multicolumn{2}{c}{Spatial} \\
\cmidrule(lr){6-7} \cmidrule(lr){8-9}
Model & Dataset &Data size & Model & Dimen. & T2T & I2T & T2T & I2T \\
\midrule 
RN50 & WebImageText & 400M & 102M & 1024 & 35.56 & 54.16 & 33.28 & 45.31 \\
RN101 & WebImageText & 400M & 120M & 512 & \underline{35.76} & 53.44 & 31.09 & 43.9 \\
CLIP-ViT-B/32 & WebImageText & 400M & 151M & 512 & 34.63 & \underline{54.88} & 30.16 & 44.69 \\
RN50x4 & WebImageText & 400M & 178M & 640 & 32.58 & 52.01 & 32.81 & \underline{45.94} \\
RN50x16 & WebImageText & 400M & 291M & 768 & 33.40 & 50.98 & \underline{34.84} & 42.66 \\
CLIP-ViT-L/14 & WebImageText & 400M & 428M & 768 & 34.94 & 52.11 & 27.19 & 42.5 \\
RN50x64 & WebImageText & 400M & 623M & 1024 & 33.40 & 52.41 & 23.13 & 40.31 \\
\midrule
RoBERTa-ViT-B/32 & LAION & 2B & 212M & 512 & 53.34 & \textbf{58.89} & \textbf{36.25} & 37.66 \\
ViT-H/14 & LAION & 2B & 986M & 1024 & 47.48 & 53.03 & 30.02 & 39.84 \\
ViT-g/14 & LAION & 2B & 1367M & 1024 & 49.23 & 55.40 & 32.5 & 40.47 \\
ViT-bigG/14 & LAION & 2B & 2540M & 1280 & 50.87 & 56.83 & 35.31 & \textbf{40.48} \\
xlm-roberta-base-ViT-B/32 & LAION & 5B & 366M & 512 & 55.09 & 52.11 & 32.66 & 37.97 \\
xlm-roberta-large-ViT-H/14 & LAION & 5B & 1193M & 1024 & \textbf{56.53} & 55.40 & 34.38 & 37.5 \\
\midrule
large:ViT-B/16 & DataComp & 1B & 150M & 512 & 35.97 & 49.43 & 22.03 & 33.91 \\
xlarge:ViT-L/14 & DataComp & 13B & 428M & 768 & \underline{46.25} & \underline{54.47} & \underline{27.97} & \underline{38.13} \\
\bottomrule
\end{tabular}
\vspace{1cm}
\end{table*}

\clearpage
\subsection{Detailed Results of VLMs on VISLA}
\label{det_res_vlm}

\textbf{Detailed results of VLMs}: In Table \ref{tab:generic}, we provide the detailed comparison of the performance of different VLMs on the generic VISLA dataset. 

\begin{table*}[htb!]
\caption{Comparison of different multi-modal vision language models performance when tested on the Generic \task benchmark (consists of 973 samples). Data Size and Model size refer to the pre-training dataset size and total number of parameters in the model (in Millions), respectively. Performance reported in terms of Accuracy (\%). P1-Ref: first positive caption compared to the negative caption; P2-Ref: second positive caption compared to the negative caption;}
\label{tab:generic}
\vspace{0.5cm}
\begin{tabular}{l|rrrcccc}
\toprule
Model & Data & Model & Emb. & P1-N & P2-N & I2T & T2T \\
\midrule 
RN50 \cite{radford2021learning} & 400M & 102M & 1024& 70.81 & 66.50 & 54.16 & 35.56 \\
RN101 \cite{radford2021learning} & 400M & 120M & 512 & 67.42 & 66.70 & 53.44 & 35.77 \\
CLIP-ViT-B-32  \cite{radford2021learning} & 400M & 151M & 512 & 68.85 & 67.63 & 54.88 & 34.64 \\
RN50x4  \cite{radford2021learning} & 400M & 178M & 640& 66.29 & 65.78 & 52.00 & 32.58 \\
RN50x16  \cite{radford2021learning} & 400M & 291M & 768 & 69.68 & 63.62 & 50.98 & 33.40 \\
CLIP-ViT-L-14  \cite{radford2021learning} & 400M & 428M & 768 & 64.13 & 67.63 & 52.11 & 34.94 \\
RN50x64 \cite{radford2021learning} & 400M & 623M & 1024  & 68.76 & 65.36 & 52.42 & 33.40 \\
\midrule
ALIGN \cite{jia2021scaling} & 1.8B & 490M & 640 & 67.01 & 62.90 & 50.46 & 42.34 \\
ALBEF \cite{li2021albef} & 14M & 210M & 256 & 76.57 & 55.09 & 49.54 & 28.26 \\
SegCLIP \cite{luo2023segclip} & 400M & 151M & 512 & 62.47 & 76.51 & 57.07  & 37.54 \\
FLAVA \cite{singh2022flava}  & 70M & 358M & 768 & 73.38 & 71.22 & 59.82  & 56.32 \\
BLIP-Caption-COCO \cite{li2022blip}  & 129M & 225M & 512 & 56.01 & 52.31 & 36.59  & 12.23 \\
BLIP-ITM-COCO \cite{li2022blip}  & 129M & 223M & 512 & 73.38 & 63.01 & 58.37  &  -- \\
BLIP2 \cite{li2023blip2}  & 129M & 1173M & 256 & 62.28 & 65.16 & 51.70 & 41.11 \\
XVLM-4M \cite{zeng2022multi} & 4M & 216M & 256 & 77.39 & 52.93 & 47.69 & 28.57 \\
XVLM-16M \cite{zeng2022multi} & 16M & 216M & 256 & 79.03 & 65.36 & 59.40 & 47.59 \\
XVLM-16M-COCO \cite{zeng2022multi} & 16M & 216M & 256 & 81.19 & 66.81 & 62.38 & 61.56 \\
XVLM-16M-Flickr \cite{zeng2022multi} & 16M & 216M & 256 & 77.70 & 68.14 & 62.08 & 58.68 \\
ViLT \cite{kim2021vilt}  & 10M & 111M & 768 & 35.97 & 76.88 & 33.61  & – \\
ViLT-ITR-COCO \cite{kim2021vilt}  & 10M & 111M & 768 & 77.18 & 67.42 & 61.05  & – \\
AltCLIP \cite{ChenLZYW23}  & 42M & 864M & 768 & 73.18 & 66.29 & 57.76  & 53.44  \\
ALIP \cite{yang2023alip}  & 15M & 151M & 512 & 74.10 & 54.47 & 47.79  & 21.07  \\
RoBERTa-ViT-B-32 \cite{schuhmann2022laion} & 2B & 212M & 512 & 71.42 & 67.83 & 58.89 & 53.34 \\
ViT-H-14 \cite{schuhmann2022laion} & 2B & 986M & 1024 & 69.58 & 62.38 & 53.03 & 47.48 \\
ViT-g-14 \cite{schuhmann2022laion} & 2B & 1367M & 1024 & 71.74 & 64.23 & 55.40 & 49.23 \\
ViT-bigG-14 \cite{schuhmann2022laion} & 2B & 2540M & 1280 & 74.62 & 64.34 & 56.84 & 50.87 \\
xlm-roberta-base-ViT-B-32 \cite{schuhmann2022laion} & 5B & 366M & 512 & 67.12 & 62.28 & 52.11 & 55.09 \\
xlm-roberta-large-ViT-H-14 \cite{schuhmann2022laion} & 5B & 1193M & 1024 & 70.92 & 63.10 & 55.40 & 56.53 \\
large:ViT-B-16 \cite{gadre2023datacomp} & 1B & 150M & 512 & 65.05 & 60.74 & 49.43 & 35.97 \\
xlarge:ViT-L-14 \cite{gadre2023datacomp} & 13B & 428M & 768 & 69.89 & 64.24 & 54.47 & 46.25 \\
\midrule
\multicolumn{5}{l}{\textbf{Models proposed to learn compositional and structural information}} \\
NegCLIP \cite{yuksekgonul2023and}  & 400M & 151M & 512 & 73.69 & 58.79 & 52.83  & 48.72 \\
CLIP-SVLC \cite{doveh2023teaching}  & 400M & 151M & 512 & 63.72 & 58.99 & 44.60 & 48.10 \\
BLIP-SGVL \cite{HerzigMKAFDG23}  & 129M & 696M & 768 & 30.52 & 29.91 & 28.88 & -- \\
CyCLIP \cite{goel2022cyclip}  & 3M & 102M & 1024 & 55.19 & 54.78 & 38.23 & 26.31 \\
\bottomrule
\end{tabular}
\end{table*}

\clearpage
\textbf{Detailed results of VLMs}: In Table \ref{tab:spatial}, we provide the detailed comparison of the performance of different VLMs on the spatial VISLA dataset.
\begin{table}[!htb]
\caption{Comparison of different multi-modal vision language models performance when tested on the spatial \task dataset (consists of 640 samples). Data Size and Model size refer to the pre-training dataset size and total number of parameters in the model (in Millions), respectively. Performance reported in terms of Accuracy (\%). P1-Ref: first positive caption compared to the negative caption; P2-Ref: second positive caption compared to the negative caption; P1-P2-Ref: first and second positive caption compared to the negative caption; Text: only text encoder considered for analysis}
\label{tab:spatial}
\begin{tabular}{l|rrrcccc}
\toprule
Model & Data & Model & Emb. & P1-N & P2-N & I2T & T2T \\
\midrule
RN50 \cite{radford2021learning} & 400M & 102M & 1024 & 64.84 & 60.15 & 45.31 & 33.28 \\
RN101 \cite{radford2021learning} & 400M & 120M & 512 & 62.81 & 57.50 & 43.90 & 31.09 \\
CLIP-ViT-B-32  \cite{radford2021learning} & 400M & 151M & 512 & 63.12 & 60.63 & 44.69 & 30.16 \\
RN50x4  \cite{radford2021learning} & 400M & 178M & 640 & 63.13 & 61.56 & 45.94 & 32.81 \\
RN50x16  \cite{radford2021learning} & 400M & 291M & 768 & 59.84 & 58.44 & 42.66 & 34.84 \\
CLIP-ViT-L-14  \cite{radford2021learning} & 400M & 428M & 768 & 62.97 & 58.75 & 42.50 & 27.19 \\
RN50x64 \cite{radford2021learning} & 400M & 623M & 1024 & 59.22 & 56.25 & 40.31 & 23.13 \\
\midrule
ALIGN \cite{jia2021scaling} & 1.8B & 490M & 640 & 53.13 & 52.98 & 35.16 & 34.53 \\
ALBEF \cite{li2021albef} & 14M & 210M & 256 & 64.38 & 60.31 & 42.66 & 25.78 \\
SegCLIP \cite{luo2023segclip} & 400M & 151M & 512 & 49.37 & 52.97 & 33.59  & 25.63 \\
FLAVA \cite{singh2022flava}  & 70M & 358M & 768 & 43.43 & 41.25 & 25.31 & 28.44 \\
BLIP-Caption-COCO \cite{li2022blip}  & 129M & 225M & 512 & 47.50 & 48.44 & 33.62  & 31.25 \\
BLIP-ITM-COCO \cite{li2022blip}  & 129M & 223M & 512 & 46.40 & 41.56 & 33.59 & – \\
BLIP2 \cite{li2023blip2}  & 129M & 1173M & 256 & 55.78 & 58.28 & 41.09 & 40.62 \\
XVLM-4M \cite{zeng2022multi} & 4M & 216M & 256 & 58.59 & 55.78 & 42.19 & 24.84 \\
XVLM-16M \cite{zeng2022multi} & 16M & 216M & 256 & 66.56 & 64.38 & 50.31 & 31.41 \\
XVLM-16M-COCO \cite{zeng2022multi} & 16M & 216M & 256 & \textbf{67.65} & \textbf{65.93} & \textbf{51.09} & \textbf{45.16} \\
XVLM-16M-Flickr \cite{zeng2022multi} & 16M & 216M & 256 & 64.53 & 60.31 & 45.16 & 39.69 \\
ViLT \cite{kim2021vilt}  & 10M & 111M & 768 & 34.06 & 48.91 & 20.32 & – \\
ViLT-ITR-COCO \cite{kim2021vilt}  & 10M & 111M & 768 & 69.06 & 62.97 & 50.16 & – \\
AltCLIP \cite{ChenLZYW23} & 42M & 864M & 768 & 63.59 & 62.19 & 45.00 & 35.16  \\
ALIP \cite{yang2023alip}  & 15M & 151M & 512 & 60.47 & 55.00 & 38.75 & 17.82 \\
RoBERTa-ViT-B-32 \cite{schuhmann2022laion} & 2B & 212M & 512 & 55.00 & 55.47 & 37.66 & 36.25 \\
ViT-H-14 \cite{schuhmann2022laion} & 2B & 986M & 1024 & 56.41 & 57.50 & 39.84 & 30.02 \\
ViT-g-14 \cite{schuhmann2022laion} & 2B & 1367M & 1024 & 57.50 & 60.31 & 40.47 & 32.50 \\
ViT-bigG-14 \cite{schuhmann2022laion} & 2B & 2540M & 1280 & 57.03 & 56.41 & 40.48 & 35.31  \\
xlm-roberta-base-ViT-B-32 \cite{schuhmann2022laion} & 5B & 366M & 512 & 57.97 & 54.84 & 37.97 & 32.66 \\
xlm-roberta-large-ViT-H-14 \cite{schuhmann2022laion} & 5B & 1193M & 1024 & 54.84 & 55.00 & 37.50 & 34.38 \\
large:ViT-B-16 \cite{gadre2023datacomp} & 1B & 150M & 512 & 56.41 & 50.31 & 33.91 & 22.03 \\
xlarge:ViT-L-14 \cite{gadre2023datacomp} & 13B & 428M & 768 & 55.63 & 55.94 & 38.13 & 27.97 \\
\midrule
\multicolumn{5}{l}{\textbf{Models proposed to learn compositional and structural information}} \\
NegCLIP \cite{yuksekgonul2023and}  & 400M & 151M & 512 & 57.34 & 54.38 & 34.84 & 29.21 \\
CLIP-SVLC \cite{doveh2023teaching}  & 400M & 151M & 512 & 59.06 & 49.22 & 28.75 & 30.94 \\
BLIP-SGVL \cite{HerzigMKAFDG23}  & 129M & 696M & 768 & 42.03 & 40.94 & 33.75 & -- \\
CyCLIP \cite{goel2022cyclip}  & 3M & 102M & 1024 & 51.10 & 48.13 & 31.41 & 12.50 \\
\bottomrule
\end{tabular}
\end{table}

\clearpage
\subsection{Qualitative Results}
\label{qual_vlm}
\subsubsection{Examples that fail on both image-to-text I2T(\xmark) and text-to-text T2T(\xmark) task}
\label{qual_vlm_1}

\begin{figure}[h!]
    \centering
    \includegraphics[width=0.82\linewidth]{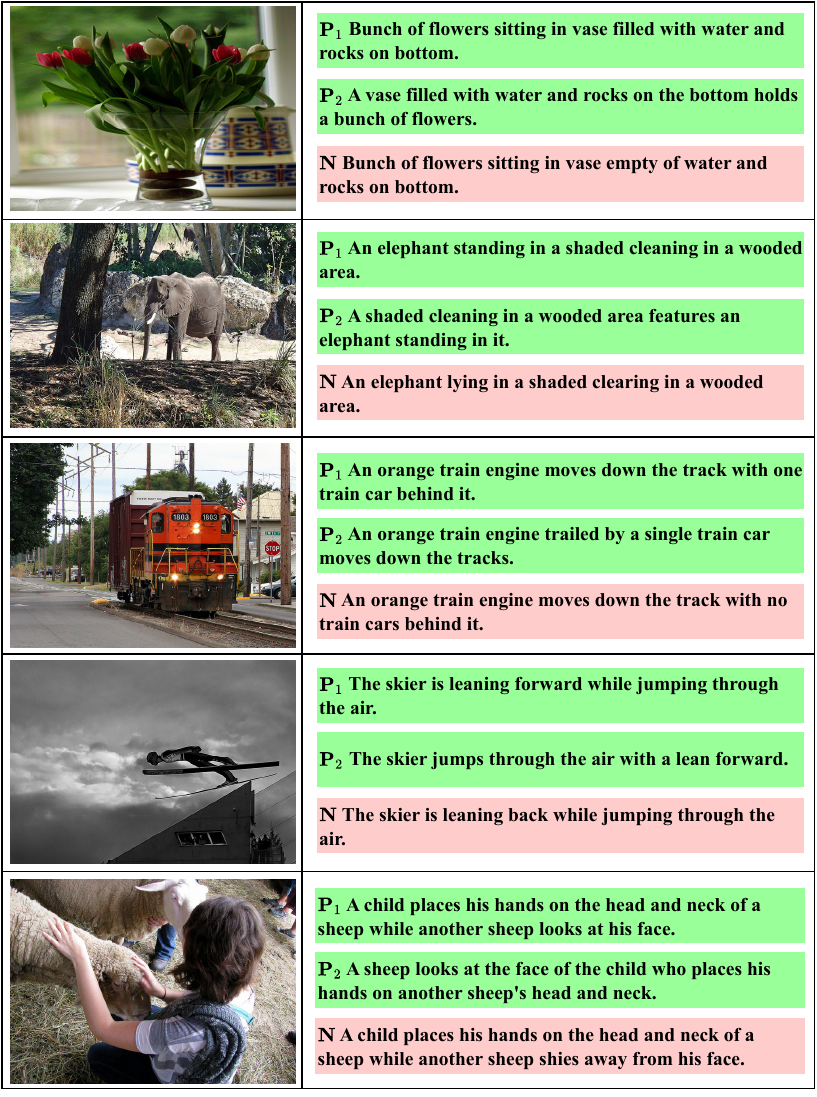}
    \caption{\small Example from the generic \task that are misclassified by VLM when both image and text are provided as input, i.e., for I2T task. These examples show that for the \task, the issues faced by the text encoders of VLMs extend to the I2T task.}
    \label{fig:err_img_txt_vlm_sc}
\end{figure}

\clearpage
\subsubsection{Examples that fail on text-to-text T2T(\xmark) and pass on image-to-text I2T(\checkmark)}
\label{qual_vlm_2}
\begin{figure}[!htb]
    \centering
    \includegraphics[width=0.79\linewidth]{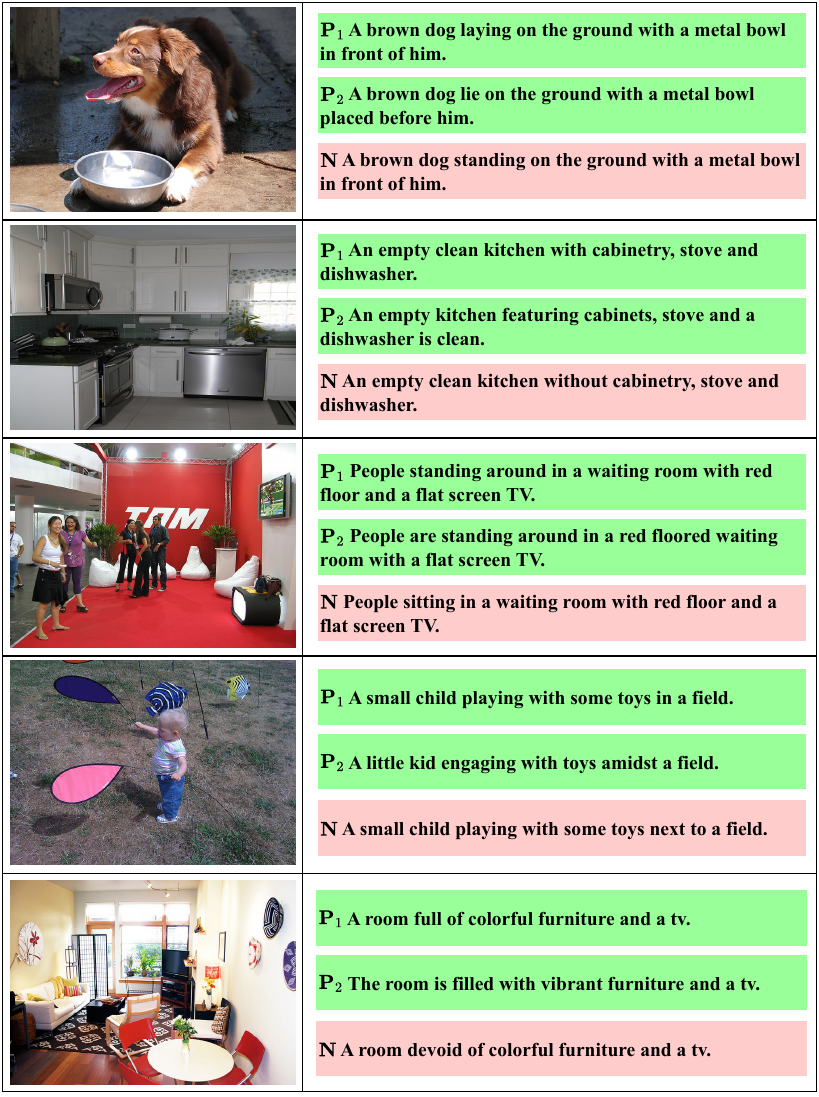}
    \caption{\small Example samples from the generic \task that are correctly recognized by the VLM when both image and text are provided as input but confused when only text input is provided (T2T task). These examples show that for the \task task, the text encoder of VLMs get confused to recognize semantically equivalent utterances even for simple lexical alterations such as negation, replacing words with synonyms, reordering of few words, etc. }
    \label{fig:img_cor_txt_error_vlm}
\end{figure}

\end{document}